\documentclass{article}
\PassOptionsToPackage{numbers, compress}{natbib}
\usepackage[main, final]{neurips_2026}
\usepackage[utf8]{inputenc}
\usepackage[T1]{fontenc}
\usepackage{hyperref}
\usepackage{url}
\usepackage{booktabs}
\usepackage{amsfonts}
\usepackage{nicefrac}
\usepackage{microtype}
\usepackage{xcolor}
\usepackage{amsmath}
\usepackage{amssymb}
\usepackage{graphicx}
\usepackage{wrapfig}
\usepackage{multirow}
\usepackage{algorithm}
\usepackage{algpseudocode}
\usepackage{subcaption} 
\usepackage{tikz}
\usepackage{pifont}

\usepackage{rotating}

\title{DyPSI: Dynamic Physics Sensing via Joint Field and Sensor-Trajectory Generation}
\author{%
\begin{tabular}[t]{c}
Yizhou Zhang$^{2}$\quad Panqi Chen$^{1}$\quad Lei Cheng$^{1}$\quad Ting Zhang$^{1}$\quad Jianlong Li$^{1}$\\
Jianfeng Shi$^{3}$\quad Shikai Fang$^{1}$\thanks{Correspond to fsk@zju.edu.cn}\\[4pt]
{\normalfont $^{1}$College of Information Science and Electronic Engineering, Zhejiang University}\\
{\normalfont $^{2}$School of Mathematical Sciences, Ocean University of China}\\
{\normalfont $^{3}$College of Energy Engineering, Zhejiang University}\\
\end{tabular}
}
\begin{document}
\maketitle
\begin{abstract}
Physics sensing in evolving environments requires reconstructing dense spatiotemporal fields from sparse observations while adapting where future observations should be collected. Existing reconstruction and sensor-placement methods have made substantial progress, but they typically rely on static sensing assumptions: they either use a single layout across all frames or optimize placements independently at each frame, which limits their ability to produce temporally coherent sensing trajectories. We introduce \emph{DyPSI}, a dynamic physics sensing framework that formulates this problem as the joint generation of future fields and sensor trajectories conditioned on historical observations. DyPSI lifts discrete sensor coordinates into continuous Sensor Position Fields, encodes fields and sensing configurations with a shared Functional Tucker representation, and trains a joint diffusion model to generate future fields and sensing trajectories simultaneously in the resulting latent space. A Gaussian-process temporal kernel correlates the perturbation injected during diffusion training, biasing the denoiser toward temporally smooth recoveries, and we further provide an analysis linking the EDM training objective to a sequence-aggregated A-optimality criterion and a kernel-induced bound on discrete trajectory variation. Experiments on turbulent flow, GLORYS12 sea-surface temperature, and 3D car aerodynamics show that DyPSI consistently outperforms static and frame-wise placement strategies, with substantial error reduction under tight sensor budgets. The code is available at \url{https://github.com/mengqi132/DyPSI}.
\end{abstract}

\section{Introduction}

Physics sensing recovers dense spatiotemporal fields from sparse sensor observations and underpins applications in fluid dynamics~\citep{hosseini2024flow, gu2007iterative, zhou2025unisolver}, climate monitoring~\citep{reynolds2007daily, reynolds2002improved}, and industrial inspection~\citep{kralovec2020review, tao2020noncontact}. In many realistic deployments, sensing resources are both sparse and \emph{mobile}: autonomous underwater vehicles~\citep{munafo2011autonomous, zhang2016autonomous}, drifting buoys~\citep{leonard2007collective, fan2025comprehensive}, and aerial platforms~\citep{burgues2020environmental, lyu2022unmanned} reposition over time to track evolving phenomena~\citep{smith2010planning}. Unlike static deployments, mobile sensing must continually reallocate observations to follow drifting vortices~\citep{taira2017modal}, propagating fronts~\citep{pascual2017multiplatform, hu2026think}, and translating surface-pressure peaks~\citep{zhao2022sparse}. Effective sensing therefore requires jointly reconstructing fields and steering sensors, since reconstruction depends on past sensor locations while informed motion depends on predicted field evolution.

Recent learning-based methods couple reconstruction with sensor-placement optimization, exemplified by PhySense~\citep{ma2025physense}, which jointly trains a reconstructor and a learnable placement under a shared loss. However, such placement either optimizes a single global layout shared across frames or optimizes each frame independently. The former cannot follow migrating structures, while the latter produces temporally discontinuous ``teleporting'' trajectories. Neither ensures that the sensor at time $t{+}1$ remains within reach of its position at time $t$, decoupling the sensing strategy from the sequential dynamics and platform kinematics.

\begin{figure}[t]
    \centering
    \includegraphics[width=\linewidth]{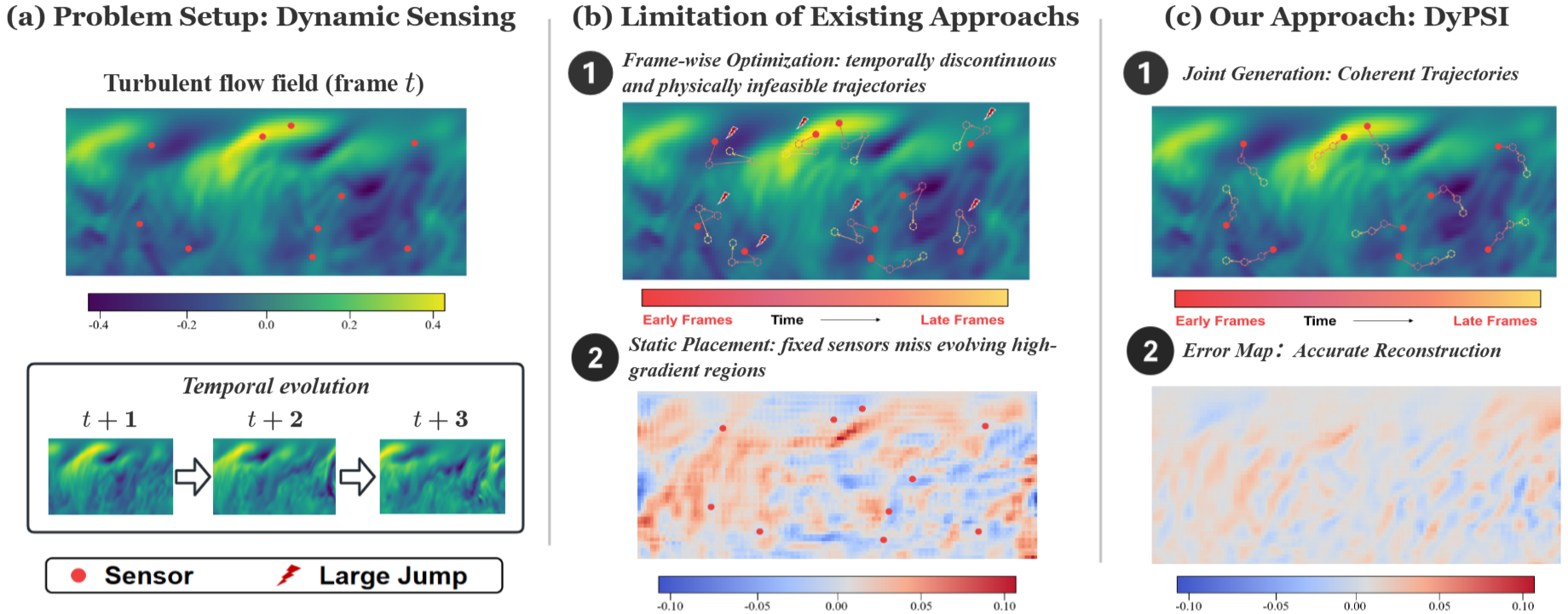}
    \caption{Dynamic active sensing. (a)~Mobile sensors (red) traverse a time-evolving turbulent flow and must continuously reposition to capture informative regions. (b)~Static placement is fixed across frames and misses migrating high-gradient structures; per-frame optimization is spatially adaptive but produces temporally discontinuous, kinematically infeasible jumps. (c)~DyPSI generates temporally coherent sensor trajectories \emph{jointly} with the reconstructed field, attaining the lowest residual error.}
    \label{fig:intro}
\end{figure}

Bridging these regimes poses three coupled challenges: a \emph{representation mismatch} between continuous high-dimensional fields and discrete coordinate sequences; a \emph{bidirectional coupling} in which future sensing depends on predicted dynamics while prediction depends on past sensing; and a \emph{temporal-consistency gap} in which per-frame optima may be globally suboptimal or kinematically infeasible. Figure~\ref{fig:intro}(a)--(b) illustrates how static layouts miss migrating structures while per-frame optimization produces jumps that real platforms cannot execute.

To address these challenges, we propose \emph{DyPSI}, a framework for joint conditional generation of future fields and sensor trajectories. We lift discrete sensor coordinates into continuous \emph{Sensor Position Fields}, encode fields and configurations in a shared Functional Tucker latent, and train a joint diffusion model to generate both simultaneously. Replacing the i.i.d.\ noise of standard EDM with Gaussian-process-correlated perturbations further encourages temporally smooth trajectories. We provide population-level analyses of sensing consistency and kinematic feasibility and validate DyPSI on three real-world physics-sensing benchmarks. As shown in Figure~\ref{fig:intro}(c), DyPSI tracks migrating informative structures with temporally coherent, reachable trajectories while accurately reconstructing the field.

\paragraph{Contributions}
\textbf{(i)~Formulation:} we recast dynamic active sensing as joint conditional generation over future fields and sensor trajectories, replacing static and per-frame placement with sequence-level joint modeling.
\textbf{(ii)~Method:} we introduce continuous Sensor Position Fields, a shared Functional Tucker latent, and a joint diffusion model with Gaussian-process-correlated perturbations for sequence-aware generation.
\textbf{(iii)~Theory and empirics:} we provide complementary sensing-consistency and kinematic-feasibility analyses and show that DyPSI consistently outperforms static and per-frame placement baselines on three real-world benchmarks.
\section{Preliminary}
\label{sec:preliminary}

This section recalls three ingredients reused throughout the paper: the Functional Tucker representation that maps fields and sensing configurations into a shared continuous-coordinate latent (\S\ref{subsec:joint_repr}), the Elucidated Diffusion Model (EDM) denoising objective that we instantiate on the joint latent (\S\ref{subsec:joint_diffusion}), and the A-optimality criterion that we extend to a sequence-aggregated form and connect to the EDM objective in \S\ref{subsec:theory_motivation}.

\subsection{Functional Tucker Representation}
\label{subsec:prelim_ftm}

For continuous coordinates $\mathbf{z}=(z_1,\ldots,z_K)\in\Omega_1\!\times\!\cdots\!\times\!\Omega_K$ and ranks $(R_1,\ldots,R_K)$, Functional Tucker representation~\citep{luo2024low, fang2024functional, li2025deep} approximates a multivariate function $f(\cdot): \mathbb{R}^{K} \rightarrow \mathbb{R} $ with Tucker decomposition~\citep{sidiropoulos2017tensor} form:
\begin{equation}
\label{eq:prelim_ftm}
f(\mathbf{z})
\;\approx\;
\sum_{r_1,\ldots,r_K}\!\mathcal{W}_{r_1,\ldots,r_K}\,
\prod_{k=1}^{K}\bigl[\boldsymbol{\phi}_k(z_k)\bigr]_{r_k}
\;=\;
\bigl\langle\operatorname{vec}(\mathcal{W}),\,
\boldsymbol{\phi}_1(z_1)\!\otimes\!\cdots\!\otimes\!\boldsymbol{\phi}_K(z_K)\bigr\rangle ,
\end{equation}
where $\mathcal{W}\in\mathbb{R}^{R_1\times\cdots\times R_K}$ is the core tensor and each $\boldsymbol{\phi}_k:\Omega_k\to\mathbb{R}^{R_k}$ is a vector-valued basis (a coordinate MLP, following SDIFT~\citep{chen2025sdift}). The Kronecker form on the right makes explicit that $f$ is a linear combination of products of single-mode basis evaluations; per-mode ranks $R_k$ are independent and can be chosen separately for each modality, enabling the shared spatial bases with modality-specific channel modes used in \S\ref{subsec:joint_repr}. 

\subsection{EDM Denoising in Latent Space}
\label{subsec:prelim_edm}

We adopt the Elucidated Diffusion Model (EDM) formulation of~\citet{karras2022elucidating}. Given a clean latent $\mathbf{x}$, a noise level $\sigma\sim p(\sigma)$, and $\mathbf{n}\sim\mathcal{N}(\mathbf{0},\mathbf{I})$, EDM trains a $\sigma$-conditioned denoiser $D_\theta$ via the squared-error objective
\begin{equation}
\label{eq:prelim_edm}
\mathcal{L}_{\mathrm{EDM}}
=
\mathbb{E}_{\mathbf{x},\sigma,\mathbf{n}}\!
\left[\lambda(\sigma)\,\bigl\|D_\theta(\mathbf{x}+\sigma\mathbf{n};\sigma)-\mathbf{x}\bigr\|_2^2\right],
\end{equation}
with noise-dependent weighting $\lambda(\sigma)$. Input/output preconditioning normalizes the denoiser's signal-to-noise ratio across $\sigma$, so a single network covers a wide range of noise scales --- a property we exploit in \S\ref{subsec:joint_diffusion} when the i.i.d.\ Gaussian draw $\mathbf{n}$ is replaced by a temporally correlated one without altering the loss form. At the population optimum, the Bayes-optimal denoiser is the conditional mean $D^{*}(\mathbf{x}+\sigma\mathbf{n};\sigma)=\mathbb{E}[\mathbf{x}\mid\mathbf{x}+\sigma\mathbf{n}]$, and its risk equals the weighted trace of the posterior covariance of $\mathbf{x}$ given the noisy observation; this identity is the lever connecting EDM training to the sequence-aggregated uncertainty criterion analysed in \S\ref{subsec:theory_motivation}.

\subsection{A-Optimality for Sensing Uncertainty}
\label{subsec:prelim_aopt}

A-optimality is the classical experimental-design criterion~\citep{pukelsheim2006optimal}: for a sensing configuration $\mathbf{s}_t$ inducing latent-field posterior covariance $\boldsymbol{\Sigma}_{\mathbf{s}_t}$, the single-step criterion is the trace of the posterior covariance,
\begin{equation}
\label{eq:prelim_aopt_single}
\mathcal{L}_A^t \;:=\; \mathrm{tr}\!\left(\boldsymbol{\Sigma}_{\mathbf{s}_t}\right).
\end{equation}
We adopt the trace criterion (rather than D-optimality's $\log\det$ or E-optimality's $\lambda_{\max}$) because it is, up to a constant, the average per-coordinate posterior variance and is therefore directly interpretable as expected squared reconstruction error --- the criterion most naturally aligned with the squared-error form of Eq.~\eqref{eq:prelim_edm}. Since dynamic sensing accumulates uncertainty over a \emph{trajectory} rather than a single layout, we work with the sequence-aggregated form
\begin{equation}
\label{eq:prelim_aopt_seq}
\mathcal{L}_A^{\mathrm{seq}}
\;:=\;
\sum_{t=1}^{T}\mathrm{tr}\!\left(\boldsymbol{\Sigma}_{\mathbf{s}_t}\right),
\end{equation}
which scores an entire trajectory by its cumulative posterior uncertainty, treating per-step posteriors independently and acting as an upper bound on any criterion that accounts for cross-time covariances. \S\ref{subsec:theory_motivation} and Appendix~\ref{app:theory} show that the EDM training risk in Eq.~\eqref{eq:prelim_edm} controls Eq.~\eqref{eq:prelim_aopt_seq} up to constant factors under standard Gaussian-posterior assumptions.

\begin{figure}[htbp]
    \centering
    \includegraphics[width=\linewidth]{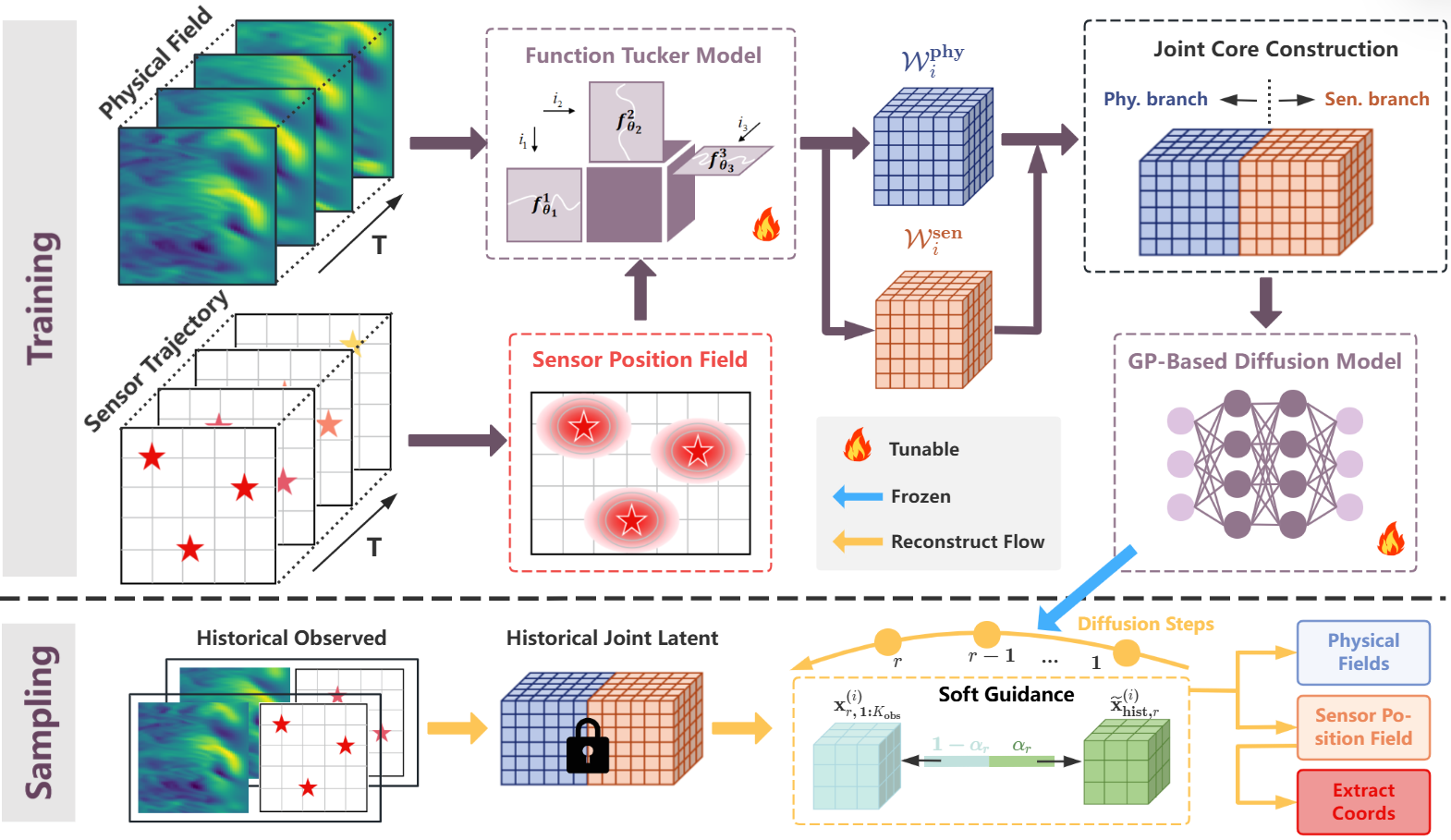}
    \caption{Overview of DyPSI. DyPSI jointly generates future physical fields and adaptive sensor trajectories via temporally correlated latent diffusion conditioned on historical observations.}
    \label{fig:overview}
\end{figure}
\section{Methodology}
\label{sec:methodology}

DyPSI generates future fields and sensor trajectories jointly from historical observations. We address the representation mismatch between physical fields and sensor trajectories with a functional Tucker latent (\S\ref{subsec:joint_repr}), build a sequence-aware joint diffusion model on top (\S\ref{subsec:joint_diffusion}), and ground both choices with two complementary population-level results (\S\ref{subsec:theory_motivation}).

\subsection{Problem Formulation}
\label{subsec:method_problem}

A physical field $\mathcal{P}_t: \Omega \to \mathbb{R}^C$ evolves over $\Omega \subset \mathbb{R}^D$ for $T$ timesteps. At each time $t$, $k$ mobile sensors with coordinates $\mathbf{s}_t=\{(x_t^i,y_t^i)\}_{i=1}^{k}\in\Omega^k$ produce noisy point observations
\begin{equation}
\label{eq:method_obs}
\mathbf{y}_t
=
\bigl[\mathcal{P}_t(x_t^1,y_t^1),\ldots,\mathcal{P}_t(x_t^k,y_t^k)\bigr]^\top+\boldsymbol{\epsilon}_t,
\qquad
\boldsymbol{\epsilon}_t\sim\mathcal{N}(\mathbf{0},\sigma_{\mathrm{obs}}^2\mathbf{I}).
\end{equation}
For notational simplicity, we use 2D sensor coordinates here, and the formulation directly extends to general $D$-dimensional domains by replacing $(x_t^i,y_t^i)$ with $\mathbf{r}_t^i\in\Omega\subset\mathbb{R}^D$.

Given the history $\mathcal{H}_t=\{(\mathbf{s}_\tau,\mathbf{y}_\tau)\}_{\tau=1}^{t}$, DyPSI models the joint conditional $p\!\left(\mathcal{P}_{t+1:T},\mathbf{s}_{t+1:T}\mid\mathcal{H}_t\right)$, in contrast to the two common placement abstractions:
\begin{equation}
\label{eq:method_three_paradigms}
\begin{aligned}
\text{Static (e.g.,\,PhySense-opt):}\quad
& \mathbf{s}^{\star}=\arg\min_{\mathbf{s}\in\Omega^k}\,\mathbb{E}_t\bigl[L(\mathcal{P}_t,\mathbf{s})\bigr], \\
\text{Per-frame:}\quad
& \mathbf{s}_t^{\star}=\arg\min_{\mathbf{s}\in\Omega^k}\,L(\mathcal{P}_t,\mathbf{s}),\quad \forall t, \\
\text{Joint (DyPSI):}\quad
& p\!\left(\mathcal{P}_{t+1:T},\,\mathbf{s}_{t+1:T}\,\bigm|\,\mathcal{H}_t\right),
\end{aligned}
\end{equation}
where $L$ is a per-frame reconstruction loss. The static formulation cannot track evolving structures; the per-frame formulation is spatially adaptive but allows $\|\mathbf{s}_{t+1}^{\star}-\mathbf{s}_t^{\star}\|$ to grow without bound, violating mobile-platform kinematics. The joint formulation places field evolution and sensor motion under a single distribution, so any temporal correlation in the model regularizes both simultaneously. It also departs from action-level control: the output is a distribution over future \emph{sequences} rather than a deterministic per-step policy.

\subsection{Joint Representation of Fields and Sensor Trajectories}
\label{subsec:joint_repr}

The two objects in Eq.~\eqref{eq:method_three_paradigms} have incompatible native types: $\mathcal{P}_t$ is a continuous high-dimensional function on $\Omega$, while $\mathbf{s}_t$ is a sparse, discrete set of coordinates. To operate on both with one generative model, we need to reconcile these types in a common space while preserving the spatial geometry they share. We do this in two steps. First, we lift the sensor configuration into a continuous surrogate on $\Omega$ via a parameter-free Gaussian density (\emph{Sensor Position Field} below). Second, we encode this surrogate together with the physical field through a Functional Tucker decomposition that shares its spatial bases between the two modalities, so that a single denoiser can attend to either modality in a single latent without architectural duplication.

\paragraph{Sensor Position Fields} For each $\mathbf{s}_t$, we define a Sensor Position Field $\mathcal{S}_t:\Omega\to\mathbb{R}_{+}$ by
\begin{equation}
\label{eq:method_spf}
\mathcal{S}_t(x,y)
=
\sum_{i=1}^{k}
\exp\!\left(
-\tfrac{(x-x_t^i)^2}{2\sigma_x^2}
-\tfrac{(y-y_t^i)^2}{2\sigma_y^2}
\right),
\end{equation}
with the analogous Gaussian-product extension in 3D. We treat the SPF as a differentiable spatial surrogate, not as a new physical variable; it lets us place sensing configurations and physical fields on the same domain so that a single Functional Tucker basis can encode both. The map is bidirectional: Eq.~\eqref{eq:method_spf} gives $\mathbf{s}_t\!\to\!\mathcal{S}_t$ in closed form, and at inference we recover $\mathbf{s}_t$ from $\mathcal{S}_t$ by a top-$k$ $\arg\max$ readout. We invoke this readout only at sampling; during training, supervision flows through the SPF latent reconstruction itself, which is fully differentiable. We defer bandwidth selection and the mesh extension to Appendix~\ref{app:impl_arch}.

\paragraph{Joint Tucker latent} We partition time into segments of length $T_{\mathrm{seg}}$, and for each segment $i$ apply Eq.~\eqref{eq:prelim_ftm} to the field and SPF sequences, producing
\begin{equation}
\label{eq:method_ftm_latents}
\mathcal{W}^{\mathrm{phy}}_{i}\!\in\!\mathbb{R}^{T_{\mathrm{seg}}\times R_1^{\mathrm{phy}}\times R_2\times\cdots\times R_{D+1}},
\quad
\mathcal{W}^{\mathrm{sen}}_{i}\!\in\!\mathbb{R}^{T_{\mathrm{seg}}\times R_1^{\mathrm{sen}}\times R_2\times\cdots\times R_{D+1}}.
\end{equation}
The encoder is \emph{amortized}: $\boldsymbol{\phi}_2,\ldots,\boldsymbol{\phi}_{D+1}$ are trained once on the entire dataset, jointly with sample-specific cores, by reconstruction against arbitrary continuous-coordinate observations of $\mathcal{P}_t$ and $\mathcal{S}_t$. Once trained, these spatial bases are \emph{shared across both modalities and across all samples and timesteps} --- justified by the fact that $\mathcal{S}_t$ and $\mathcal{P}_t$ live on the same $\Omega$ --- with only the channel-mode rank $R_1$ differing between $\mathcal{P}$ and $\mathcal{S}$. The core tensors of Eq.~\eqref{eq:method_ftm_latents} therefore carry all sample- and time-specific information, and become the only objects on which the diffusion model operates. We concatenate them along the channel mode,
\begin{equation}
\label{eq:method_joint_core}
\mathbf{x}_{\mathrm{joint}}^{(i)}
=
\mathrm{Concat}_{R_1}\!\bigl[\mathcal{W}^{\mathrm{phy}}_{i},\,\mathcal{W}^{\mathrm{sen}}_{i}\bigr]
\in\mathbb{R}^{T_{\mathrm{seg}}\times R_{\mathrm{joint}}\times R_2\times\cdots\times R_{D+1}},
\quad R_{\mathrm{joint}}=R_1^{\mathrm{phy}}+R_1^{\mathrm{sen}},
\end{equation}
so that fields and sensing become a single sequence-valued latent on which one denoiser can attend to either modality at every layer.

\subsection{Joint Latent Diffusion: Training and Sampling}
\label{subsec:joint_diffusion}

We use a conditional denoiser 
$D_\theta(\,\cdot\,;\sigma,\mathbf{h}_{\mathrm{hist}}^{(i)})$ 
that takes a $\sigma$-noised joint latent and the encoded history 
$\mathbf{h}_{\mathrm{hist}}^{(i)}$ as input, and predicts the clean joint latent 
$\mathbf{x}_{\mathrm{joint}}^{(i)}$. During training, $D_\theta$ is trained on joint trajectories constructed by a costly offline sequence-level optimizer. At test time, $D_\theta$ generates future fields and sensor trajectories conditioned only on historical observations. Details of training trajectory construction and network architecture are provided in Appendices~\ref{app:impl_arch} and~\ref{app:data_prep}.

\paragraph{From frame-independent to sequence-aware noise} To produce temporally coherent trajectories, we follow the functional diffusion of sequence data ~\citet{chen2025sdift,bilos2023modeling,kerrigan2023diffusion}: instead of the i.i.d.\ Gaussian noise of standard EDM (Eq.~\eqref{eq:prelim_edm}), we corrupt the latent with a single Gaussian-Process(GP) sample $\boldsymbol{\eta}^{(i)}$ correlated along the time mode of $\mathbf{x}_{\mathrm{joint}}^{(i)}$. This re-frames the problem as functional temporal diffusion over sequence-valued latents rather than pointwise diffusion over per-frame noise. The motivation is direct: under i.i.d.\ noise, each frame in $\mathbf{x}_{\mathrm{joint}}^{(i)}$ is corrupted independently and the denoiser recovers each frame in isolation, leaving no handle on trajectory smoothness at sampling time. A GP-correlated draw instead couples adjacent frames in the denoising task, making the resulting denoiser sequence-aware --- without adding any regularizer to the loss.

\paragraph{Mat\'ern-3/2 correlated perturbations} We instantiate the correlated draw with a Mat\'ern-$3/2$ kernel. Let $\mathbf{K}_{\mathrm{time}}\in\mathbb{R}^{T_{\mathrm{seg}}\times T_{\mathrm{seg}}}$ denote its Gram matrix on the segment with Cholesky factor $\mathbf{L}\mathbf{L}^{\top}=\mathbf{K}_{\mathrm{time}}$, and let $\boldsymbol{\xi}^{(i)}\sim\mathcal{N}(\mathbf{0},\mathbf{I})$ be a standard Gaussian tensor of the same shape as $\mathbf{x}_{\mathrm{joint}}^{(i)}$. We then define
\begin{equation}
\label{eq:method_corr_noise}
\boldsymbol{\eta}^{(i)}
=
\bigl(\mathbf{L}\otimes \mathbf{I}_{R_{\mathrm{joint}}R_2\cdots R_{D+1}}\bigr)\,\operatorname{vec}(\boldsymbol{\xi}^{(i)}),
\end{equation}
reshaped back to the latent shape. By construction $\boldsymbol{\eta}^{(i)}$ is correlated only along the time mode and remains independent on all other modes; the kernel itself corresponds to a once-differentiable Gauss--Markov prior, and we discuss the rationale for this choice over RBF and white-noise endpoints in Appendix~\ref{app:impl_train}. By forcing $D_\theta$ to recover $\mathbf{x}_{\mathrm{joint}}^{(i)}$ from a temporally smooth perturbation, we bias its outputs toward temporally smooth trajectories --- both for fields and for sensor positions --- which is the regularity we want for mobile sensing. We train with the EDM loss but with $\boldsymbol{\eta}^{(i)}$ in place of $\mathbf{n}$:
\begin{equation}
\label{eq:method_joint_diffusion}
\mathcal{L}_{\mathrm{joint}}
=
\mathbb{E}_{i,\sigma,\boldsymbol{\xi}}\!
\left[\lambda(\sigma)\,
\bigl\|D_\theta(\mathbf{x}_{\mathrm{joint}}^{(i)}+\sigma\boldsymbol{\eta}^{(i)};\sigma,\mathbf{h}_{\mathrm{hist}}^{(i)})-\mathbf{x}_{\mathrm{joint}}^{(i)}\bigr\|_2^2\right].
\end{equation}

\paragraph{Conditional posterior sampling} At inference, we observe the first $K_{\mathrm{obs}}$ frames of a segment and must generate the remainder. We split $\mathbf{x}^{(i)}=(\mathbf{x}^{(i)}_{1:K_{\mathrm{obs}}},\mathbf{x}^{(i)}_{K_{\mathrm{obs}}+1:T_{\mathrm{seg}}})$ and softly anchor the observed prefix during reverse denoising:
\begin{equation}
\label{eq:method_guidance}
\mathbf{x}_{r,\,1:K_{\mathrm{obs}}}^{(i)}
\leftarrow
(1-\alpha_r)\,\mathbf{x}_{r,\,1:K_{\mathrm{obs}}}^{(i)}+\alpha_r\,\widetilde{\mathbf{x}}_{\mathrm{hist},r}^{(i)},
\end{equation}
where $\widetilde{\mathbf{x}}_{\mathrm{hist},r}^{(i)}$ is the historical latent perturbed to the current noise level and $\alpha_r$ anneals from $1$ to $0$ across reverse steps (schedule in Appendix~\ref{app:impl_sample}). The suffix follows the standard reverse dynamics. After sampling, we decode both modalities by inverse Functional Tucker and recover sensor coordinates from the SPF channel via the top-$k$ $\arg\max$ readout of \S\ref{subsec:joint_repr}
\subsection{Theoretical Properties}
\label{subsec:theory_motivation}

The analysis below provides two population-level consistency checks for the design of \S\ref{subsec:joint_diffusion}. First, under an idealized linear-Gaussian posterior model, a denoising/reconstruction risk is aligned with a sequence-aggregated A-optimality criterion, motivating the reconstruction-based trajectory supervision used by the offline optimizer. Second, the GP kernel used for temporal perturbations induces an algebraic bound on the discrete variation of trajectories favored by the corresponding negative log-prior. The first result is derived for the standard EDM loss with i.i.d.\ noise; the correlated perturbation in Eq.~\eqref{eq:method_corr_noise} is an additional temporal-coherence design and is not used in the closed-form Bayes-risk derivation.

\paragraph{Sensing consistency.}
Active sensing is fundamentally about uncertainty reduction: an informative trajectory should shrink the posterior covariance of the field. The next proposition shows that, under standard linear-Gaussian assumptions, an idealized EDM denoising risk is equivalent up to constants to a sequence-aggregated posterior-variance criterion.

\textbf{Proposition 1 (EDM risk and sequence A-optimality).} \textit{Under Assumptions A.1--A.3 of Appendix~\ref{app:theory}, there exist constants $C_{\mathrm{low}},C_{\mathrm{high}}>0$ such that}
\begin{equation}
\label{eq:theory_prop1}
C_{\mathrm{low}}\sum_{t=1}^{T}\mathrm{tr}(\boldsymbol{\Sigma}_{\mathbf{s}_t})
\;\le\;
\mathcal{L}_{\mathrm{EDM},*}^{\mathrm{seq}}
\;\le\;
C_{\mathrm{high}}\sum_{t=1}^{T}\mathrm{tr}(\boldsymbol{\Sigma}_{\mathbf{s}_t}).
\end{equation}
\textit{This result analyzes an idealized denoising posterior conditioned on sparse observations, and therefore supports the reconstruction-based trajectory-supervision objective rather than proving that the finite-sample joint diffusion loss directly optimizes A-optimality.}

\paragraph{Kinematic feasibility.}
The Mat\'ern-$3/2$ correlated noise of Eq.~\eqref{eq:method_corr_noise} is the only sequence-aware ingredient in DyPSI's diffusion loss; we do \emph{not} add an explicit GP regularizer during diffusion training. Nevertheless, the same kernel defines a GP negative log-prior $\mathcal{L}_{\mathrm{GP}}(\mathbf{s})=\tfrac{1}{2}\mathbf{s}^{\top}\mathbf{K}^{-1}\mathbf{s}$ on trajectories. The next proposition characterizes the kinematic content of this kernel-level prior.

\textbf{Proposition 2 (GP regularization controls trajectory variation).}
\textit{Let $\mathbf{D}$ be the first-order time-difference matrix and 
$c_{\mathbf{K}}:=\lambda_{\max}(\mathbf{D}\mathbf{K}\mathbf{D}^{\top})$ 
for a positive-definite GP kernel matrix $\mathbf{K}$. For a zero-mean or time-constant-reference GP prior,}
\begin{equation}
\label{eq:theory_prop2}
\sum_{t=1}^{T-1}\bigl\|\mathbf{s}_{t+1}-\mathbf{s}_t\bigr\|_2^2
\;\le\;
2\,c_{\mathbf{K}}\,\mathcal{L}_{\mathrm{GP}}(\mathbf{s}).
\end{equation}
\textit{For a time-varying reference mean, an additional reference-variation term appears; see Appendix~\ref{app:theory}. The result is a kernel-level algebraic characterization, not a control-theoretic guarantee.}

Together, the two propositions motivate the two design axes of \S\ref{subsec:joint_diffusion}: reconstruction-based trajectory supervision aligns with posterior-variance reduction in an idealized setting, while the temporal GP kernel biases generated trajectories toward smooth motion. Full assumptions and proofs are in Appendix~\ref{app:theory}.
\section{Related Work}
\label{sec:related_work}

\paragraph{Sparse field reconstruction.}
Reconstructing dense spatiotemporal fields from sparse observations is a classical ill-posed inverse problem~\citep{kabanikhin2011inverse}. Deterministic deep models such as VoronoiCNN~\citep{fukami2021global} and Senseiver~\citep{santos2023senseiver} recover dense fields from scattered measurements through spatial interpolation structures or attention-based latent representations. Generative approaches including DiffusionPDE~\citep{huang2024diffusionpde}, S$^3$GM~\citep{li2024learning}, and diffusion posterior sampling~\citep{chung2023diffusion} instead model conditional field distributions under partial observation, and SDIFT~\citep{chen2025sdift} generates full-field evolution from irregular sparse observations in Functional Tucker space. In all of these methods the sensing locations are fixed, random, externally given, or optimized outside the generative model, whereas DyPSI generates future fields and future sensing trajectories jointly. A detailed itemized delineation between DyPSI and SDIFT is given in Appendix~\ref{app:sdift_delineation}.

\paragraph{Sensor placement and experimental design.}
Sensor placement is rooted in optimal experimental design~\citep{papalambros2000principles, pukelsheim2006optimal}, where A-, D-, and E-optimality criteria quantify the information content of a sensing configuration. Classical data-driven methods exploit low-dimensional structure through POD and QR based sensor selection~\citep{manohar2018data}, compressed sensing~\citep{donoho2006compressed}, and Gaussian process placement~\citep{krause2008near}. PhySense~\citep{ma2025physense} couples learning-based reconstruction with placement optimization and is closest to our work, but its layouts remain static or frame-wise independent, while DyPSI treats field evolution and sensor motion as one sequence-level generative problem.

\paragraph{Dynamic and active sensing.}
Informative path planning, model predictive control, and reinforcement learning address adaptive sensing by optimizing action sequences against explicit reward or information objectives under a known or learned dynamics model~\citep{krause2008near, singh2009efficient, binney2012branch, morere2017sequential}. These methods require a specified reward and per-step decision procedure and typically plan over actions alone, with reconstruction handled by a separate estimator. DyPSI addresses the complementary formulation of jointly generating future fields and sensing trajectories conditioned on history, amortizing sequence-level planning into a single conditional sampling procedure. The analyses in \S\ref{subsec:theory_motivation} serve as design-time consistency checks rather than control-theoretic guarantees.
\section{Experiments}
\label{sec:experiments}

We evaluate DyPSI on three physics sensing tasks: two-dimensional turbulent channel flow, global sea surface temperature recovery with geographic masking, and surface pressure reconstruction over an irregular three-dimensional car mesh. These benchmarks cover regular Cartesian grids, constrained geospatial domains, and unstructured surface geometries in 2D and 3D settings.

\textbf{Benchmarks:} The benchmarks include: (1) Turbulent flow: two-dimensional channel-flow turbulence on a $128\times48$ regular grid~\citep{santos2023senseiver}; (2) Sea temperature: daily global ocean surface temperature at $1^\circ$ resolution ($360\times180$) from the GLORYS12 reanalysis dataset spanning 27 years~\citep{lellouche2021glorys12}; (3) Car aerodynamics: surface pressure fields simulated via OpenFOAM~\citep{jasak2009openfoam} for a passenger-car mesh with 95,428 vertices from ShapeNet~\citep{chang2015shapenet}.

\textbf{Baselines and Experimental Settings:} We compare against SSPOR~\citep{manohar2018data}, VoronoiCNN~\citep{fukami2021global}, Senseiver~\citep{santos2023senseiver}, S$^3$GM~\citep{li2024learning}, DiffusionPDE~\citep{huang2024diffusionpde}, and PhySense~\citep{ma2025physense} under both random layouts and static optimized placement. Identical training budgets are enforced across all methods, and reconstruction quality is measured by the relative $L_2$ metric. For turbulent flow, during training, sensor counts from 25 to 300 are randomly instantiated with coordinates sampled uniformly across the spatial extent, and the reconstructor is assessed under varying sensor layouts. Because DiffusionPDE and S$^3$GM incur prohibitive inference costs, their assessment is restricted to 500 test cases. For sea temperature, fields are down-sampled to $1^\circ\times1^\circ$ resolution. During training, 10 to 100 sensors are randomly selected from valid ocean pixels, with continental land masses masked out. The training corpus comprises 9,843 daily snapshots from 1993 to 2019, and the test period covers 2020 to 2021. A binary land mask is supplied as auxiliary input to all methods. For car aerodynamics, aerodynamic pressure distributions are obtained via OpenFOAM simulations across inlet velocities of 20 to 40 m/s and yaw angles of $-10^\circ$ to $10^\circ$. Sensors are restricted to the vehicle exterior. Unless otherwise stated, DyPSI uses PhySense as the default teacher reconstructor for offline trajectory-supervision construction; all DyPSI test errors are computed from fields generated by the trained DyPSI model, not from test-time oracle optimization. All experiments are conducted with five independent random seeds. The main comparison tables report mean values for compactness, while Appendix~\ref{app:full_results} provides the complete results with standard deviations and paired $t$-test $p$-values against the strongest baseline in each setting. Further implementation details are relegated to Appendix~\ref{app:impl}.

\begin{figure}[t]
\centering
\begin{minipage}[t]{0.26\textwidth}
\vspace{0pt}
\makebox[0pt][l]{\small\textbf{(a) Kinematic Trajectory Evolution}}
\end{minipage}%
\hfill
\begin{minipage}[t]{0.73\textwidth}
\vspace{0pt}
\centering
\small\textbf{(b) Case Study: Turbulent Flow}
\end{minipage}

\vspace{0.2em}

\begin{subfigure}[t]{0.24\textwidth}
\centering
\vspace{0pt}
\includegraphics[width=\linewidth]{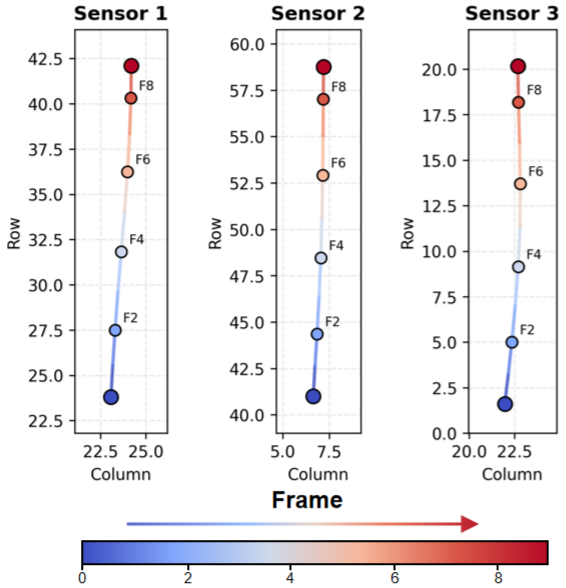}
\end{subfigure}%
\hfill
\begin{subfigure}[t]{0.75\textwidth}
\centering
\vspace{0pt}
\begin{tikzpicture}
\node[anchor=south west,inner sep=0] (image) at (0,0) {\includegraphics[width=\linewidth]{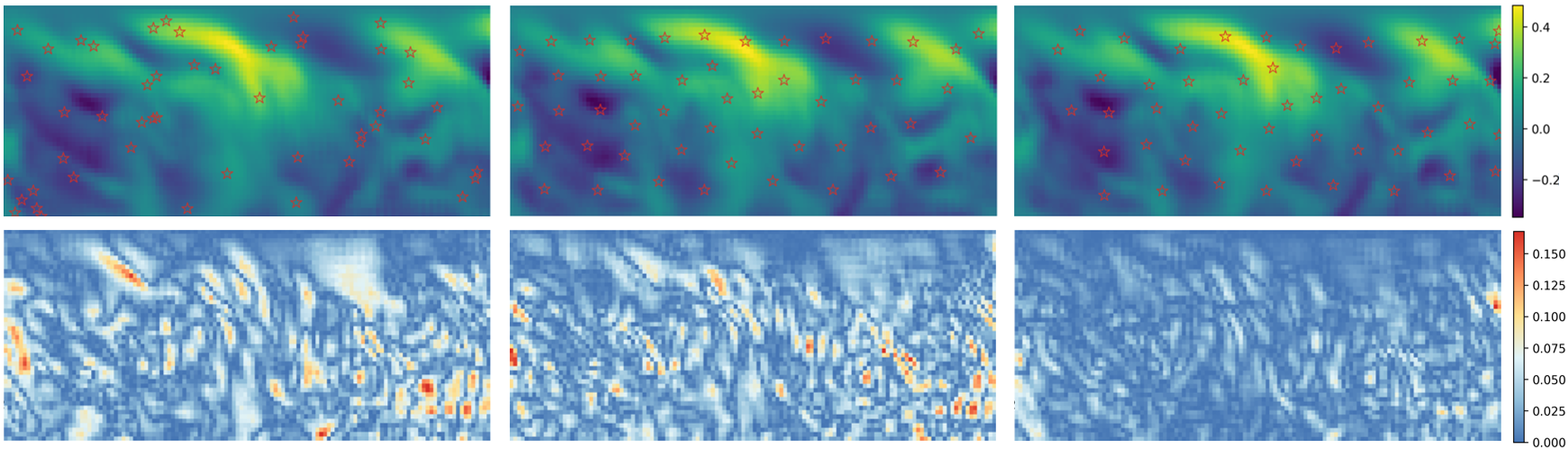}};
\begin{scope}[x={(image.south east)},y={(image.north west)}]
\node[anchor=north,font=\footnotesize] at (0.14,-0.03) {Random Sensor Placement};
\node[anchor=north,font=\footnotesize] at (0.48,-0.03) {PhySense Optimized Placement};
\node[anchor=north,font=\footnotesize] at (0.82,-0.03) {DyPSI Optimized Placement};
\end{scope}
\end{tikzpicture}
\end{subfigure}

\caption{Qualitative and kinematic analysis on the turbulent flow benchmark. (a) Temporally coherent trajectories of 3 mobile sensors generated by DyPSI across frames 0-9. Color encodes temporal progression from early (blue) to late (red). (b) Reconstructed fields (top row) and point-wise error maps (bottom row) under random, static, and dynamic 50-sensor placements (denoted by \ding{73}). }
\label{fig:turbulent}

\end{figure}

\begin{table}[t]
\centering
\footnotesize
\setlength{\tabcolsep}{5.5pt}
\caption{Performance comparison on the turbulent flow and sea surface temperature benchmarks. Relative $L_2$ error is reported. DyPSI uses dynamically generated sensor trajectories; unless otherwise specified, its offline trajectory supervision is constructed using PhySense as the default teacher reconstructor. DyPSI errors are computed from fields generated by the trained joint diffusion model, without test-time oracle placement. \# denotes the sensor count. Improvement quantifies the relative error reduction with respect to the strongest static baseline.}

\label{tab:main_2d}
\begin{tabular}{ll|cccc|cccc}
\toprule
 &  & \multicolumn{4}{c|}{Turbulent Flow} & \multicolumn{4}{c}{Sea Temperature} \\
Models & Placement & \#200 & \#100 & \#50 & \#30 & \#100 & \#50 & \#25 & \#15 \\
\midrule
SSPOR & SSPOR & 0.3018 & 0.4429 & 0.5921 & 0.6637 & 0.0756 & 0.0752 & 0.0789 & 0.0719 \\
VoronoiCNN & Random & 0.3588 & 0.4870 & 0.6919 & 0.7992 & 0.1496 & 0.1537 & 0.2433 & 0.2817 \\
Senseiver & Random & 0.1842 & 0.2316 & 0.3740 & 0.5746 & 0.0715 & 0.0732 & 0.0769 & 0.0784 \\
S$^3$GM & Random & 0.1856 & 0.2016 & 0.2681 & 0.6421 & 0.1115 & 0.1137 & 0.1581 & 0.1795 \\
DiffusionPDE & Random & 0.1993 & 0.3312 & 0.6604 & 0.9163 & 0.0664 & 0.0775 & 0.0859 & 0.1139 \\
PhySense & Random & 0.1233 & 0.1527 & 0.2586 & 0.5176 & 0.0439 & 0.0452 & 0.0477 & 0.0520 \\
PhySense-opt & Static Opt. & 0.1106 & 0.1257 & 0.1558 & 0.2157 & 0.0426 & 0.0430 & 0.0437 & 0.0439 \\
\textbf{DyPSI} & Dynamic Gen. & \textbf{0.0986} & \textbf{0.1101} & \textbf{0.1352} & \textbf{0.1826} & \textbf{0.0387} & \textbf{0.0390} & \textbf{0.0394} & \textbf{0.0402} \\
\midrule
\multicolumn{2}{l|}{Improvement} & 10.87\% & 12.41\% & 13.19\% & 15.35\% & 9.15\% & 9.30\% & 9.84\% & 8.43\% \\
\bottomrule
\end{tabular}
\vspace{-12pt}
\end{table}

\begin{wraptable}{r}{0.52\textwidth}
\centering
\footnotesize
\setlength{\tabcolsep}{3pt}
\caption{Results on the car aerodynamics benchmark.}
\label{tab:car}
\begin{tabular}{l|ccccc}
\toprule
Models & \#200 & \#100 & \#50 & \#30 & \#15 \\
\midrule
Senseiver & 0.1009 & 0.1012 & 0.1018 & 0.1022 & 0.1053 \\
DiffusionPDE & 0.0967 & 0.0966 & 0.0987 & 0.1055 & 0.2095 \\
PhySense & 0.0375 & 0.0382 & 0.0395 & 0.0416 & 0.0465 \\
PhySense-opt & 0.0369 & 0.0370 & 0.0370 & 0.0372 & 0.0386 \\
\textbf{DyPSI} & \textbf{0.0301} & \textbf{0.0301} & \textbf{0.0302} & \textbf{0.0307} & \textbf{0.0311} \\
\midrule
Improvement & 18.43\% & 18.65\% & 18.39\% & 17.47\% & 19.43\% \\
\bottomrule
\end{tabular}
\vspace{-12pt}
\end{wraptable}

Tables~\ref{tab:main_2d} and~\ref{tab:car} show that DyPSI achieves the lowest reconstruction error across all benchmarks and sensor budgets. Relative to PhySense-opt, it reduces error by 10.87\% to 15.35\% on turbulent flow, 8.43\% to 9.84\% on sea surface temperature, and 17.47\% to 19.43\% on car aerodynamics. The gains generally become more pronounced as sensor availability decreases, indicating that temporal adaptation is especially valuable when observations are sparse. On the car mesh, 15 dynamically coordinated sensors attain an error of 0.0311, outperforming PhySense-opt with 30 sensors and PhySense with 100 randomly placed sensors. Appendix~\ref{app:full_results} further shows that all turbulent flow and car aerodynamics improvements are significant at $p<0.001$, while the sea surface temperature improvements are significant at $p\leq0.01$. The consistency across regular grids, masked geospatial domains, and unstructured surface meshes indicates that the benefit is not restricted to a particular discretization. All DyPSI errors are produced by the learned generator without test-time oracle placement.

Figure~\ref{fig:turbulent} visualizes three smooth trajectories that follow coherent flow structures over frames 0 to 9 without abrupt positional jumps. Under the 50-sensor setting, random placement leaves distributed residuals, while static optimization retains errors near structures that have advected away from the fixed sensors. DyPSI reduces these localized errors by repositioning sensors around informative, high-gradient filaments. The sea temperature trajectories similarly respect the land mask and concentrate sensing within thermally informative ocean regions.

\begin{table}[t]
\centering
\scriptsize
\setlength{\tabcolsep}{2.4pt}
\caption{Accuracy and kinematic feasibility under representative sparse sensor budgets. Mean values over five independent runs are reported. Viol. denotes the velocity violation rate, Smooth. denotes $\sum_t\|\mathbf{s}_{t+1}-2\mathbf{s}_t+\mathbf{s}_{t-1}\|^2$, and runtime is measured in seconds per 64-frame sequence. The Per-frame Oracle and Sequence Teacher access future ground truth and perform test-time optimization, whereas DyPSI uses neither.}
\label{tab:kinematic_tradeoff}
\resizebox{\textwidth}{!}{%
\begin{tabular}{lcc|cccc|cccc}
\toprule
 &  &  & \multicolumn{4}{c|}{\textbf{Turbulent Flow, 50 sensors}} & \multicolumn{4}{c}{\textbf{Sea Surface Temperature, 25 sensors}} \\
Strategy & Future GT & Test-time Opt. & Rel. $L_2$ $\downarrow$ & Viol. $\downarrow$ & Smooth. $\downarrow$ & Runtime $\downarrow$ & Rel. $L_2$ $\downarrow$ & Viol. $\downarrow$ & Smooth. $\downarrow$ & Runtime $\downarrow$ \\
\midrule
Static PhySense-opt & $\times$ & $\times$ & 0.1558 & 0 & 0 & $-$ & 0.0437 & 0 & 0 & $-$ \\
Per-frame Oracle & $\checkmark$ & $\checkmark$ & 0.1007 & 89\% & 142.5 & 887 & 0.0259 & 75\% & 165.3 & 628 \\
Sequence Teacher & $\checkmark$ & $\checkmark$ & 0.1206 & 5\% & 3.5 & 1683 & 0.0355 & 2\% & 4.5 & 1429 \\
\textbf{DyPSI} & $\times$ & $\times$ & 0.1352 & 8\% & 4.2 & \textbf{1.9} & 0.0394 & 4\% & 9.2 & \textbf{2.1} \\
\bottomrule
\end{tabular}%
}
\end{table}

\textbf{Accuracy and Kinematic Feasibility:} Table~\ref{tab:kinematic_tradeoff} compares four sensing strategies on identical test sequences with a frozen reconstructor. The Per-frame Oracle independently applies PGD at each future frame without temporal or kinematic constraints, whereas the Sequence Teacher optimizes complete trajectories using future ground truth. DyPSI instead generates fields and trajectories from historical observations without future ground truth or test-time optimization.

The Per-frame Oracle obtains the lowest errors of 0.1007 on turbulent flow and 0.0259 on sea surface temperature, but violates the velocity limit in 89\% and 75\% of transitions. Its smoothness costs of 142.5 and 165.3 also reveal severe inter-frame discontinuities. DyPSI reduces the violation rates to 8\% and 4\% and the smoothness costs to 4.2 and 9.2, while remaining within 0.0146 and 0.0039 of the Sequence Teacher in reconstruction error. It requires only 1.9 and 2.1 seconds per sequence, compared with 628 to 1683 seconds for the optimization-based alternatives. These results quantify the accuracy and computational trade-offs of adaptive sensing under kinematic constraints. DyPSI thus provides a practical intermediate between non-adaptive static placement and costly ground-truth-informed trajectory optimization.

\begin{table}[t]
\centering
\footnotesize
\setlength{\tabcolsep}{2.2pt}
\caption{Constructor/backbone generalization on the turbulent flow benchmark. Relative $L_2$ error is reported. For non-DyPSI rows, column groups denote the reconstruction backbone used for evaluation. For the DyPSI-opt row, column groups denote the teacher reconstructor used by the offline trajectory-supervision constructor to train DyPSI; the reported error is computed from fields generated by the trained DyPSI model.}
\label{tab:placement_generalization}
\begin{tabular}{l|cccc|cccc|cccc}
\toprule
 & \multicolumn{4}{c|}{Senseiver} & \multicolumn{4}{c|}{PhySense} & \multicolumn{4}{c}{DiffusionPDE} \\
Placement Strategy & \#200 & \#100 & \#50 & \#30 & \#200 & \#100 & \#50 & \#30 & \#200 & \#100 & \#50 & \#30 \\
\midrule
Random Sample & 0.1842 & 0.2316 & 0.3740 & 0.5746 & 0.1233 & 0.1527 & 0.2586 & 0.5176 & 0.1993 & 0.3312 & 0.6604 & 0.9163 \\
Grid Sample & 0.1615 & 0.2273 & 0.3845 & 0.5869 & 0.1235 & 0.1521 & 0.2436 & 0.4930 & 0.1932 & 0.3319 & 0.6703 & 0.9079 \\
Min-Max Sample & 0.1549 & 0.2091 & 0.3795 & 0.5747 & 0.1196 & 0.1533 & 0.3157 & 0.5022 & 0.1889 & 0.3307 & 0.6574 & 0.9010 \\
PhySense-opt & 0.1388 & 0.1750 & 0.3025 & 0.4570 & 0.1106 & 0.1257 & 0.1558 & 0.2157 & 0.1672 & 0.2841 & 0.5319 & 0.7904 \\
\textbf{DyPSI-opt} & \textbf{0.1259} & \textbf{0.1670} & \textbf{0.2863} & \textbf{0.4527} & \textbf{0.0986} & \textbf{0.1101} & \textbf{0.1352} & \textbf{0.1826} & \textbf{0.1569} & \textbf{0.2640} & \textbf{0.5028} & \textbf{0.7116} \\
\bottomrule
\end{tabular}
\end{table}

\begin{wraptable}{r}{0.43\textwidth}
\centering
\footnotesize
\setlength{\tabcolsep}{4pt}
\caption{GP-kernel ablation on turbulent flow with 50 sensors. Mean $\pm$ standard deviation over five runs is reported.}
\label{tab:gp_kernel_ablation}
\begin{tabular}{l|cc}
\toprule
Noise type & Rel. $L_2$ $\downarrow$ & Smoothness $\downarrow$ \\
\midrule
i.i.d. & 0.1515 $\pm$ 0.0039 & 18.6 $\pm$ 1.9 \\
RBF & 0.1489 $\pm$ 0.0025 & 6.2 $\pm$ 1.3 \\
\textbf{Matérn-3/2} & \textbf{0.1352 $\pm$ 0.0024} & \textbf{4.2 $\pm$ 1.0} \\
\bottomrule
\end{tabular}
\vspace{-10pt}
\end{wraptable}

\textbf{Constructor and Backbone Generalization:} Table~\ref{tab:placement_generalization} evaluates conventional placements with three reconstruction backbones and uses each backbone independently as the teacher for a separately trained DyPSI model. No multi-teacher training or test-time ground-truth optimization is used, and the PhySense teacher setting matches the default configuration in the main comparison. Grid and Min-Max sampling do not consistently improve over random placement, indicating that spatial regularity alone is insufficient for tracking informative structures. DyPSI-opt achieves the lowest error for every teacher and sensor budget. With 30 sensors, it reduces the error from 0.4570 to 0.4527 for Senseiver, from 0.2157 to 0.1826 for PhySense, and from 0.7904 to 0.7116 for DiffusionPDE. Although the margin varies with backbone capacity, the consistent gains across implicit, flow-based, and diffusion-based teachers show that the learned dynamic sensing strategy is not tied to a particular reconstruction architecture.

\begin{wraptable}{r}{0.43\textwidth}
\centering
\footnotesize
\setlength{\tabcolsep}{4pt}
\caption{Joint and separate generation on turbulent flow with 50 sensors. Mean $\pm$ standard deviation over five runs is reported.}
\label{tab:joint_generation_ablation}
\begin{tabular}{l|cc}
\toprule
Model & Rel. $L_2$ $\downarrow$ & Smoothness $\downarrow$ \\
\midrule
Separate models & 0.1421 $\pm$ 0.0045 & 9.1 $\pm$ 1.1 \\
\textbf{Joint model} & \textbf{0.1352 $\pm$ 0.0024} & \textbf{4.2 $\pm$ 1.0} \\
\bottomrule
\end{tabular}
\vspace{-10pt}
\end{wraptable}

\textbf{Sensitivity to the GP Kernel:} We replace the Matérn-3/2 temporal perturbation with i.i.d. EDM noise or an RBF kernel on turbulent flow with 50 sensors. As shown in Table~\ref{tab:gp_kernel_ablation}, i.i.d. noise increases both reconstruction error and trajectory irregularity. RBF improves smoothness relative to i.i.d. noise but retains a higher reconstruction error, suggesting that its smooth prior restricts necessary temporal variation. Matérn-3/2 achieves the best accuracy and smoothness, supporting its balance between temporal coherence and trajectory flexibility.

We further compare joint generation with separate diffusion models that share only the history encoder. As shown in Table~\ref{tab:joint_generation_ablation}, joint generation reduces the relative $L_2$ error from 0.1421 to 0.1352 and the smoothness cost from 9.1 to 4.2. The simultaneous gains indicate that modeling both modalities within a shared denoiser better captures the dependence between field evolution and sensor motion.

\section{Limitations}
\label{sec:limitations}

Several scope boundaries qualify our results. First, training trajectories are constructed by an offline optimizer with access to ground-truth fields, so DyPSI inherits a teacher dependence; the gain-retention analysis in \S\ref{sec:experiments} quantifies this amortization gap, and DyPSI does not fully match its oracle teacher. Second, like all diffusion-based reconstructors, DyPSI assumes deployment dynamics remain within the support of the training distribution; under abrupt regime shifts or strongly non-Gaussian sensing noise the generated fields and trajectories may be unreliable, and we make no out-of-distribution claims. Third, our evaluation is a history-conditioned open-loop rollout rather than fully closed-loop sensing, although the low inference cost permits receding-horizon re-planning. Fourth, our feasibility guarantees cover kinematic and domain constraints only, namely velocity and acceleration limits, land masking, and surface projection, and do not model full platform dynamics such as underactuation or ocean currents. Fifth, at sensor densities far above those evaluated here, SPF peaks may merge and the top-$k$ readout requires the non-maximum-suppression fallback described in Appendix~C. Finally, we assume point-wise observations, whereas real sensors such as radar often integrate over small spatial neighborhoods; incorporating regional observation models into both reconstruction and trajectory generation is a natural extension.
\section{Conclusion}
\label{sec:conclusion}

We present DyPSI, a dynamic physics sensing framework that jointly generates future physical fields and adaptive sensor trajectories by encoding both modalities into a shared Functional Tucker latent space via continuous Sensor Position Fields, where a Gaussian-process temporal kernel encourages temporal coherence during diffusion training. Complementary theoretical analyses provide design-time consistency checks, linking an idealized denoising risk to sequence-aggregated posterior variance and characterizing the trajectory-variation bias induced by the GP kernel.  Extensive experiments on turbulent flow, sea-surface temperature, and car aerodynamics demonstrate that DyPSI consistently outperforms static and per-frame placement strategies, achieving particularly pronounced gains under sparse sensor budgets.
\newpage
\begin{ack}
This work was supported in part by the National Natural Science Foundation of China under Grant 6260070841 and in part by Zhejiang University Education Foundation Qizhen Scholar Foundation.
\end{ack}
\bibliographystyle{plainnat}
\bibliography{main}
\appendix
\section{Theoretical Details}
\label{app:theory}

This appendix provides the detailed assumptions and proof sketches for the theoretical motivation in the main paper. The results are established at the population level. They characterize idealized risks induced by simplified probabilistic models, rather than the approximation or optimization error of a finite-capacity neural network trained with finite data. Therefore, the theory should be interpreted as design-time consistency checks for the proposed objectives and kernels, rather than as a proof of global optimality for the practical DyPSI training procedure. Proposition A.6 analyzes the standard Gaussian denoising form of the EDM objective; the temporally correlated noise used in the full method is an additional design for temporal coherence, rather than part of the closed-form Bayes-risk derivation below.

\paragraph{Notation.}
Let $\mathbf{x}_t \in \mathbb{R}^d$ denote the latent representation of the target physical field at timestep $t$, and let $\mathbf{s}_t$ denote the corresponding sensor configuration. Given $\mathbf{s}_t$ and the associated noisy sparse measurements $\mathbf{y}_t$, the posterior uncertainty of $\mathbf{x}_t$ is described by a covariance matrix $\boldsymbol{\Sigma}_{\mathbf{s}_t} \in \mathbb{R}^{d \times d}$.

\textbf{Assumption A.1 (Conditional Gaussian Posterior).}
\textit{For each timestep $t$, conditioned on the sensor configuration $\mathbf{s}_t$ and the corresponding noisy sparse measurements $\mathbf{y}_t$, the latent field posterior is Gaussian:}
\begin{equation}
\mathbf{x}_t \mid (\mathbf{s}_t, \mathbf{y}_t)
\sim
\mathcal{N}(\boldsymbol{\mu}_{\mathbf{s}_t}, \boldsymbol{\Sigma}_{\mathbf{s}_t}),
\end{equation}
\textit{where $\boldsymbol{\Sigma}_{\mathbf{s}_t}$ is positive semidefinite.}

\textbf{Assumption A.2 (Uniform Spectral Bound).}
\textit{There exists a finite constant $\Lambda_{\max} > 0$ such that}
\begin{equation}
\lambda_{\max}(\boldsymbol{\Sigma}_{\mathbf{s}_t}) \le \Lambda_{\max},
\qquad
\forall t \in \{1,\dots,T\}, \ \forall \mathbf{s}_t \in \mathcal{D},
\end{equation}
\textit{where $\mathcal{D}$ denotes the feasible set of sensor placements.}

\textbf{Assumption A.3 (Admissible EDM Weighting).}
\textit{The EDM noise distribution $p(\sigma)$ and weighting function $\lambda(\sigma)$ satisfy}
\begin{equation}
\mathbb{E}_{\sigma \sim p(\sigma)}[\lambda(\sigma)] < \infty,
\qquad
\mathbb{E}_{\sigma \sim p(\sigma)}
\left[
\lambda(\sigma)\frac{\sigma^2}{\Lambda_{\max}+\sigma^2}
\right] > 0.
\end{equation}

\textbf{Definition A.4 (Population-Optimal EDM Risk).}
\textit{For a fixed timestep $t$, define the population-optimal EDM objective as}
\begin{equation}
\mathcal{L}_{\mathrm{EDM},*}^t(\mathbf{s}_t)
:=
\mathbb{E}_{\sigma \sim p(\sigma)}
\left[
\lambda(\sigma)
\inf_{D}
\mathbb{E}_{\mathbf{x}_t,\mathbf{n}}
\left[
\|D(\mathbf{x}_t + \sigma \mathbf{n}; \sigma) - \mathbf{x}_t\|_2^2
\right]
\right],
\end{equation}
\textit{where $\mathbf{n} \sim \mathcal{N}(\mathbf{0}, \mathbf{I})$, and the infimum is taken over all measurable denoisers $D$. The sequence-level risk is}
\begin{equation}
\mathcal{L}_{\mathrm{EDM},*}^{\mathrm{seq}}
:=
\sum_{t=1}^T \mathcal{L}_{\mathrm{EDM},*}^t.
\end{equation}

\textbf{Definition A.5 (Sequence-Aggregated A-Optimality).}
\textit{For each timestep $t$, define}
\begin{equation}
\mathcal{L}_A^t := \mathrm{tr}(\boldsymbol{\Sigma}_{\mathbf{s}_t}),
\end{equation}
\textit{and the corresponding sequence-aggregated criterion}
\begin{equation}
\mathcal{L}_A^{\mathrm{seq}}
:=
\sum_{t=1}^T \mathcal{L}_A^t
=
\sum_{t=1}^T \mathrm{tr}(\boldsymbol{\Sigma}_{\mathbf{s}_t}).
\end{equation}
\textit{This is a temporal aggregation of classical A-optimality, rather than a full joint spatiotemporal covariance criterion.}

\textbf{Proposition A.6 (EDM Objective Linearly Controls Sequence-Aggregated A-Optimality).}
\textit{Under Assumptions A.1--A.3, there exist constants $C_{\mathrm{low}}, C_{\mathrm{high}} > 0$ such that}
\begin{equation}
C_{\mathrm{low}} \, \mathcal{L}_A^{\mathrm{seq}}
\le
\mathcal{L}_{\mathrm{EDM},*}^{\mathrm{seq}}
\le
C_{\mathrm{high}} \, \mathcal{L}_A^{\mathrm{seq}},
\end{equation}
\textit{where}
\begin{equation}
C_{\mathrm{high}}
=
\mathbb{E}_{\sigma \sim p(\sigma)}[\lambda(\sigma)],
\qquad
C_{\mathrm{low}}
=
\mathbb{E}_{\sigma \sim p(\sigma)}
\left[
\lambda(\sigma)\frac{\sigma^2}{\Lambda_{\max}+\sigma^2}
\right].
\end{equation}

\textit{Proof sketch.}
Fix a timestep $t$ and a noise level $\sigma$. Under Assumption A.1, the denoising problem
\begin{equation}
\mathbf{y} = \mathbf{x}_t + \sigma \mathbf{n},
\qquad
\mathbf{n} \sim \mathcal{N}(\mathbf{0}, \mathbf{I}),
\end{equation}
is a linear Gaussian estimation problem. Hence, the Bayes-optimal denoiser is the conditional mean
\begin{equation}
D^*(\mathbf{y};\sigma) = \mathbb{E}[\mathbf{x}_t \mid \mathbf{y}],
\end{equation}
and its minimum mean-square error equals the trace of the posterior covariance. Let $\{\lambda_{i,t}\}_{i=1}^d$ denote the eigenvalues of $\boldsymbol{\Sigma}_{\mathbf{s}_t}$. By spectral decomposition, the optimal single-step EDM risk is
\begin{equation}
\mathcal{L}_{\mathrm{EDM},*}^t
=
\mathbb{E}_{\sigma \sim p(\sigma)}
\left[
\lambda(\sigma)
\sum_{i=1}^d
\frac{\lambda_{i,t}\sigma^2}{\lambda_{i,t}+\sigma^2}
\right].
\end{equation}
For the upper bound, since $\sigma^2/(\lambda_{i,t}+\sigma^2) \le 1$, we obtain
\begin{equation}
\mathcal{L}_{\mathrm{EDM},*}^t
\le
\mathbb{E}_{\sigma}[\lambda(\sigma)]
\sum_{i=1}^d \lambda_{i,t}
=
C_{\mathrm{high}} \, \mathcal{L}_A^t.
\end{equation}
For the lower bound, Assumption A.2 implies $\lambda_{i,t} \le \Lambda_{\max}$, hence
\begin{equation}
\frac{\sigma^2}{\lambda_{i,t}+\sigma^2}
\ge
\frac{\sigma^2}{\Lambda_{\max}+\sigma^2},
\end{equation}
which yields
\begin{equation}
\mathcal{L}_{\mathrm{EDM},*}^t
\ge
\mathbb{E}_{\sigma}
\left[
\lambda(\sigma)\frac{\sigma^2}{\Lambda_{\max}+\sigma^2}
\right]
\sum_{i=1}^d \lambda_{i,t}
=
C_{\mathrm{low}} \, \mathcal{L}_A^t.
\end{equation}
Summing over $t=1,\dots,T$ proves the result. \hfill $\square$

\paragraph{Interpretation.}
Proposition A.6 should be read as an idealized posterior-denoising result. Under a conditional Gaussian posterior model, the optimal denoising risk is equivalent up to constant factors to a sequence-aggregated posterior-variance criterion. This motivates the reconstruction-based trajectory-supervision objective used by the offline optimizer.
\medskip

To characterize the temporal regularity induced by the GP kernel, we consider the corresponding Gaussian-process negative log-prior on sensor trajectories.

\textbf{Definition A.7 (GP Temporal Regularizer).}
\textit{Let $\mathbf{s}^{(j,d)} \in \mathbb{R}^{T}$ denote coordinate $d$ of the trajectory of sensor $j$ over the discrete time grid. For a positive definite temporal kernel matrix $\mathbf{K}$ and mean vector $\mathbf{m}^{(j,d)}$, define}
\begin{equation}
\mathcal{L}_{\mathrm{GP}}(\mathbf{s}^{(j,d)})
:=
\frac{1}{2}
(\mathbf{s}^{(j,d)} - \mathbf{m}^{(j,d)})^\top
\mathbf{K}^{-1}
(\mathbf{s}^{(j,d)} - \mathbf{m}^{(j,d)}).
\end{equation}
\textit{For a $D$-dimensional trajectory with $k$ sensors, the total GP regularizer is the coordinate-wise sum}
\begin{equation}
\mathcal{L}_{\mathrm{GP}}(\mathbf{s})
:=
\sum_{j=1}^{k}\sum_{d=1}^{D}
\mathcal{L}_{\mathrm{GP}}(\mathbf{s}^{(j,d)}).
\end{equation}

\textbf{Proposition A.8 (GP Regularization Controls Discrete Trajectory Variation).}
\textit{Let $\mathbf{D} \in \mathbb{R}^{(T-1)\times T}$ be the first-order difference matrix defined by}
\begin{equation}
(\mathbf{D}\mathbf{u})_t = \mathbf{u}_{t+1} - \mathbf{u}_t.
\end{equation}
\textit{Define}
\begin{equation}
c_{\mathbf{K}} := \lambda_{\max}(\mathbf{D}\mathbf{K}\mathbf{D}^\top).
\end{equation}
\textit{Then, for any trajectory coordinate $\mathbf{s}^{(j,d)}$ and reference mean $\mathbf{m}^{(j,d)}$,}
\begin{equation}
\sum_{t=1}^{T-1}
\left[
\left(s_{t+1}^{(j,d)} - m_{t+1}^{(j,d)}\right)
-
\left(s_t^{(j,d)} - m_t^{(j,d)}\right)
\right]^2
\le
2 c_{\mathbf{K}} \, \mathcal{L}_{\mathrm{GP}}(\mathbf{s}^{(j,d)}).
\end{equation}
\textit{Thus the GP penalty controls the discrete variation of the trajectory relative to its reference mean. For the absolute variation,}
\begin{equation}
\sum_{t=1}^{T-1}
\left(
s_{t+1}^{(j,d)} - s_t^{(j,d)}
\right)^2
\le
4 c_{\mathbf{K}} \, \mathcal{L}_{\mathrm{GP}}(\mathbf{s}^{(j,d)})
+
2
\sum_{t=1}^{T-1}
\left(
m_{t+1}^{(j,d)} - m_t^{(j,d)}
\right)^2 .
\end{equation}
\textit{If the reference mean is zero or constant in time, the sharper bound}
\begin{equation}
\sum_{t=1}^{T-1}
\left(
s_{t+1}^{(j,d)} - s_t^{(j,d)}
\right)^2
\le
2 c_{\mathbf{K}} \, \mathcal{L}_{\mathrm{GP}}(\mathbf{s}^{(j,d)})
\end{equation}
\textit{holds. Consequently, for $k$ sensors in $D$ spatial dimensions,}
\begin{equation}
\sum_{t=1}^{T-1}
\sum_{j=1}^{k}
\left\|
\mathbf{s}_{t+1}^{(j)}-\mathbf{s}_{t}^{(j)}
\right\|_2^2
\le
4 c_{\mathbf{K}} \, \mathcal{L}_{\mathrm{GP}}(\mathbf{s})
+
2
\sum_{t=1}^{T-1}
\sum_{j=1}^{k}
\left\|
\mathbf{m}_{t+1}^{(j)}-\mathbf{m}_{t}^{(j)}
\right\|_2^2 .
\end{equation}
\textit{For zero-mean or time-constant-reference GP priors, this reduces to}
\begin{equation}
\sum_{t=1}^{T-1}
\sum_{j=1}^{k}
\left\|
\mathbf{s}_{t+1}^{(j)}-\mathbf{s}_{t}^{(j)}
\right\|_2^2
\le
2 c_{\mathbf{K}} \, \mathcal{L}_{\mathrm{GP}}(\mathbf{s}).
\end{equation}
\textit{The corresponding discrete kinetic-energy form follows by dividing both sides by $(\Delta t)^2$.}

\textit{Proof sketch.}
Let $\mathbf{u}^{(j,d)} := \mathbf{s}^{(j,d)} - \mathbf{m}^{(j,d)}$. By definition,
\begin{equation}
2\mathcal{L}_{\mathrm{GP}}(\mathbf{s}^{(j,d)})
=
(\mathbf{u}^{(j,d)})^\top
\mathbf{K}^{-1}
\mathbf{u}^{(j,d)}.
\end{equation}
The discrete variation relative to the reference mean is
\begin{equation}
\sum_{t=1}^{T-1}
\left(
u_{t+1}^{(j,d)} - u_t^{(j,d)}
\right)^2
=
(\mathbf{u}^{(j,d)})^\top
\mathbf{D}^\top\mathbf{D}
\mathbf{u}^{(j,d)}.
\end{equation}
By a generalized Rayleigh quotient argument,
\begin{equation}
(\mathbf{u}^{(j,d)})^\top
\mathbf{D}^\top\mathbf{D}
\mathbf{u}^{(j,d)}
\le
\lambda_{\max}
\!\left(
\mathbf{K}^{1/2}\mathbf{D}^\top\mathbf{D}\mathbf{K}^{1/2}
\right)
(\mathbf{u}^{(j,d)})^\top
\mathbf{K}^{-1}
\mathbf{u}^{(j,d)}.
\end{equation}
Since
\begin{equation}
\lambda_{\max}
\!\left(
\mathbf{K}^{1/2}\mathbf{D}^\top\mathbf{D}\mathbf{K}^{1/2}
\right)
=
\lambda_{\max}(\mathbf{D}\mathbf{K}\mathbf{D}^\top)
=
c_{\mathbf{K}},
\end{equation}
we obtain
\begin{equation}
\sum_{t=1}^{T-1}
\left(
u_{t+1}^{(j,d)} - u_t^{(j,d)}
\right)^2
\le
2c_{\mathbf{K}}
\mathcal{L}_{\mathrm{GP}}(\mathbf{s}^{(j,d)}).
\end{equation}
This proves the relative-variation bound. For a general time-varying mean,
\begin{equation}
s_{t+1}^{(j,d)}-s_t^{(j,d)}
=
\left(u_{t+1}^{(j,d)}-u_t^{(j,d)}\right)
+
\left(m_{t+1}^{(j,d)}-m_t^{(j,d)}\right),
\end{equation}
and applying $(a+b)^2\le 2a^2+2b^2$ gives the stated general bound. If the reference mean is zero or constant in time, the second term vanishes and the sharper bound follows directly without this relaxation. Summing over coordinates and sensors proves the vector-valued statement. \hfill $\square$

\paragraph{Interpretation.}
Proposition A.8 shows that the GP negative log-prior controls the discrete variation of a trajectory relative to its prescribed mean. When the GP mean is zero or constant in time, this directly yields a bound on absolute frame-to-frame sensor motion; for a time-varying mean, the absolute motion is additionally controlled by the variation of the reference mean. This is a kernel-level characterization of temporal smoothness, not a control-theoretic guarantee for the deployed generative sampler.

\section{Delineation from SDIFT}
\label{app:sdift_delineation}

DyPSI builds on two ingredients introduced by SDIFT~\citep{chen2025sdift}: the Functional Tucker representation used to encode fields into a continuous-coordinate latent, and the use of Gaussian-process-correlated perturbations within an EDM diffusion objective. These components are inherited and are not claimed as new in isolation. The task formulation, the generated object, and all sensing-specific machinery of DyPSI are new. Table~\ref{tab:sdift_delineation} summarizes the relationship item by item, and we expand on the key distinctions below.

\begin{table}[h]
\centering
\caption{Itemized delineation between SDIFT and DyPSI. Inherited components are explicitly credited; the remaining rows are new in this work.}
\label{tab:sdift_delineation}
\renewcommand{\arraystretch}{1.25}
\resizebox{\textwidth}{!}{
\begin{tabular}{@{}lll@{}}
\toprule
\textbf{Aspect} & \textbf{SDIFT~\citep{chen2025sdift}} & \textbf{DyPSI (Ours)} \\
\midrule
Task & Full-field generation from irregular sparse observations & Dynamic active sensing: coupled field reconstruction and adaptive path planning \\
Generated object & Physical fields $\mathcal{P}_{t+1:T}$ only & Joint fields $\mathcal{P}_{t+1:T}$ and sensor trajectories $\mathbf{s}_{t+1:T}$ \\
Sensor role & Fixed, exogenous conditioning variable & A jointly generated modality \\
Sensor representation & None & Sensor Position Field: fixed-kernel encoding of discrete coordinates into a continuous surrogate \\
Latent structure & Functional Tucker over fields & Joint core with shared spatial bases, modality-specific channel ranks, and a concatenated unified latent \\
Training supervision & Observed fields & Offline-to-online amortization of a GT-informed sequence-level optimizer \\
Temporal treatment & GP-correlated noise & GP-correlated noise plus a kinematic analysis linking the kernel to bounded trajectory variation \\
Theory & --- & EDM risk linked to sequence-aggregated A-optimality (Proposition~1) \\
Evaluation & Field reconstruction benchmarks & Dynamic sensing with kinematic-feasibility metrics and oracle trade-off analysis \\
\bottomrule
\end{tabular}}
\end{table}

SDIFT addresses reconstruction: given irregular sparse observations, it generates the evolution of the physical field, with sensor locations supplied externally and held fixed as conditioning. DyPSI addresses dynamic active sensing, in which sensor placement is itself an output of the model. The diffusion model in DyPSI generates future fields and future sensor trajectories jointly from the observed history, so the sensing configuration is lifted from a conditioning variable to a first-class generated modality. This changes both the modeling target and the evaluation: generated fields are scored against held-out ground truth, and generated trajectories are additionally scored by kinematic-feasibility metrics that have no counterpart in SDIFT.

Because sensor coordinates are discrete and low-dimensional while physical fields are continuous functions on the spatial domain, the two cannot be encoded by a shared representation directly. DyPSI introduces the Sensor Position Field, a deterministic fixed-kernel encoding that lifts each sensor configuration into a continuous surrogate on the same domain as the field. The encoding is closed form and fully differentiable in both directions, with coordinates recovered by a top-$k$ argmax readout that is exact whenever the kernel bandwidth is below the typical inter-sensor spacing, a condition that holds at all sensor densities evaluated in this work. SDIFT contains no such mechanism, since it never needs to generate sensor positions.

DyPSI retains the Functional Tucker backbone but restructures the latent: spatial bases are shared across the physical-field and SPF branches, channel-mode ranks are modality-specific, and the two cores are concatenated into a single sequence-valued joint latent on which one denoiser attends to both modalities at every layer. The ablation in \S\ref{sec:experiments} shows that this joint construction outperforms two independent diffusion models coupled only through the history encoder on both reconstruction accuracy and trajectory smoothness. Training supervision also departs from SDIFT: rather than fitting observed fields, DyPSI amortizes a costly offline sequence-level optimizer that has access to ground-truth fields, distilling its dynamic sensing strategy into a history-conditioned generator that requires neither future ground truth nor test-time optimization. The gain-retention analysis in \S\ref{sec:experiments} quantifies how much of the teacher's advantage survives this amortization.

In short, SDIFT generates fields \emph{given} sensors, while DyPSI generates sensors \emph{together with} fields. The Sensor Position Field, the joint latent, the offline-to-online amortization of trajectory supervision, the kinematic analysis, and the dynamic active-sensing evaluation are new contributions of this work. Figure~2 reuses the encode--decode schematic of SDIFT only for the shared representation stage, as noted in its caption.
\section{Implementation Details}
\label{app:impl}

This appendix collects the metrics, architecture, training, sampling, and per-benchmark hyperparameter choices used in \S\ref{sec:experiments}, together with detailed rationales that were referenced from but omitted in the main text. Processed data and the full repository will be released upon acceptance; an anonymized code repository is available at \url{https://github.com/mengqi132/DyPSI}.

\subsection{Metrics}
\label{app:metrics}
To evaluate the accuracy of the reconstructed physical field, we employ the widely used \emph{relative L2} metric, defined as
\begin{equation}
\text{Relative } L_2 = \frac{\|\mathbf{u} - \hat{\mathbf{u}}\|_2}{\|\mathbf{u}\|_2},
\end{equation}
where $\mathbf{u}$ denotes the reference (ground-truth) field and $\hat{\mathbf{u}}$ represents the model reconstruction.

\subsection{Architecture}
\label{app:impl_arch}

\paragraph{Functional Tucker basis networks.} Each per-mode basis $\boldsymbol{\phi}_k:\Omega_k\to\mathbb{R}^{R_k}$ is parameterized by a 4-layer coordinate MLP of width 128 with sinusoidal positional encoding. For 2D benchmarks the spatial bases share weights across the two coordinate axes; for the 3D mesh benchmark each axis uses an independent MLP, and the surface-mesh embedding is precomputed via the signed-distance projection of the original ShapeNet vertices.

\paragraph{Joint denoiser.} The denoiser $D_\theta$ is a U-Net operating on $\mathbf{x}_{\mathrm{joint}}^{(i)}\in\mathbb{R}^{T_{\mathrm{seg}}\times R_{\mathrm{joint}}\times R_2\times\cdots\times R_{D+1}}$. It uses standard EDM preconditioning~\citep{karras2022elucidating} with skip- and residual-scalings $c_{\mathrm{skip}}, c_{\mathrm{out}}, c_{\mathrm{in}}, c_{\mathrm{noise}}$ as defined therein, treats the modality channel $R_{\mathrm{joint}}$ as feature dimension, and incorporates temporal self-attention along the $T_{\mathrm{seg}}$ axis at each resolution.

\paragraph{History encoder.} The segment-level history encoding $\mathbf{h}_{\mathrm{hist}}^{(i)}$ is produced by a 4-layer Transformer over the observed prefix $(\mathbf{s}_{\tau},\mathbf{y}_{\tau})_{\tau=1}^{K_{\mathrm{obs}}}$ embedded by a shared MLP. The encoded vector is broadcast and concatenated to the denoiser feature map at every block.

\paragraph{SPF bandwidth selection (extended).} Equation~\eqref{eq:method_spf} requires choosing $(\sigma_x,\sigma_y)$ (and $\sigma_z$ for $D=3$). We pick $\sigma_x=\sigma_y\!\approx\!1.5\Delta$, where $\Delta$ is the local grid spacing, so that the $3\sigma$ support of each Gaussian blob covers roughly $9\!\times\!9$ grid points --- locally supported but, at the typical sensor densities of our benchmarks, with negligible overlap. Empirically, varying $\sigma$ within a factor of two has no noticeable effect on reconstruction error. For the surface-mesh case we use a geodesic-distance-based kernel with the same effective support measured in mesh edge lengths; per-benchmark values are listed in Appendix~\ref{app:impl_hp}.

\subsection{Training}
\label{app:impl_train}

We train all benchmarks with Adam ($\mathrm{lr}=2{\times}10^{-4}$, $\beta_1{=}0.9$, $\beta_2{=}0.999$), batch size 16 segments, and 200k iterations. The EDM noise schedule follows~\citet{karras2022elucidating}: $\ln\sigma\sim\mathcal{N}(P_{\mathrm{mean}}{=}{-}1.2,P_{\mathrm{std}}{=}1.2)$ with weighting $\lambda(\sigma)=(\sigma^2{+}\sigma_{\mathrm{data}}^2)/(\sigma\cdot\sigma_{\mathrm{data}})^2$ and $\sigma_{\mathrm{data}}{=}0.5$. Segment length is $T_{\mathrm{seg}}{=}16$. Mixed-precision (bfloat16) is used throughout.

\paragraph{Mat\'ern-3/2 kernel choice (extended).} The temporal kernel $\mathbf{K}_{\mathrm{time}}$ in Eq.~\eqref{eq:method_corr_noise} uses signal variance $\sigma_{\mathrm{GP}}^2{=}1$ and length-scale $\ell{=}4$ frames, so that the effective correlation horizon spans roughly a quarter of one training segment. We chose Mat\'ern-$3/2$ over the two natural endpoints --- the squared-exponential (RBF) kernel and white noise --- for the following reason. Mat\'ern-$3/2$ corresponds to a once-differentiable Gauss--Markov prior whose 1D spectral density decays as $O(\omega^{-4})$, matching the ``locally smooth, long-range independent'' regime of typical mobile-sensing trajectories. By contrast, the $C^{\infty}$ RBF kernel imposes infinitely many derivatives of regularity and tends to over-smooth legitimate inter-frame attention shifts, while white noise reduces to the standard EDM perturbation and forfeits sequence awareness altogether. Mat\'ern-$3/2$ sits between these extremes and was chosen accordingly.

\begin{table}[h]
\centering
\footnotesize
\caption{Benchmark-specific hyperparameters for the experiments of \S\ref{sec:experiments}.}
\label{tab:hp}
\begin{tabular}{l|ccc}
\toprule
& Turbulent Flow & Sea Temperature & Car Aerodynamics \\
\midrule
Spatial dimension $D$              & 2 & 2 & 3 \\
Grid / mesh size                   & $128\times48$ & $360\times180$ & 95{,}428 vertices \\
Channels $C$                       & 2 (velocity)   & 1 (SST)        & 1 (pressure) \\
Modality channel ranks $(R_1^{\mathrm{phy}}, R_1^{\mathrm{sen}})$ & $(16, 8)$ & $(16, 8)$ & $(24, 8)$ \\
Spatial ranks $(R_2,\dots,R_{D+1})$ & $(32, 16)$ & $(48, 32)$ & $(24, 24, 16)$ \\
SPF bandwidth (in grid units)      & $(1.5, 1.5)$ & $(1.5, 1.5)$ & $(2.0, 2.0, 2.0)$ \\
Segment length $T_{\mathrm{seg}}$  & 16 & 16 & 16 \\
Observed prefix $K_{\mathrm{obs}}$ & 4  & 4  & 4 \\
Sensor budget $k$ (test-time)      & $\{30,50,100,200\}$ & $\{15,25,50,100\}$ & $\{15,30,50,100,200\}$ \\
\bottomrule
\end{tabular}
\end{table}

\subsection{Sampling}
\label{app:impl_sample}

At inference we use the second-order Heun sampler from~\citet{karras2022elucidating} with 50 reverse steps. The soft-guidance coefficient $\alpha_r$ in Eq.~\eqref{eq:method_guidance} is annealed linearly from $\alpha_0{=}1$ at the highest noise level to $\alpha_R{=}0$ at the cleanest noise level, ensuring the prefix is exactly anchored at the start of denoising and freely refined toward the end. Sensor coordinates are extracted from each generated SPF $\widehat{\mathcal{S}}_t$ by a top-$k$ $\arg\max$ readout: we take the $k$ grid points (or mesh vertices, in 3D) with the largest SPF response.

\paragraph{Why naive top-$k$ $\arg\max$ suffices.} Because the bandwidth in Eq.~\eqref{eq:method_spf} is set narrower than the typical inter-sensor spacing (\S\ref{app:impl_arch}), each generated sensor contributes a well-separated peak, and the magnitude-based readout returns one location per sensor without any post-processing. We verified this empirically by inspecting the histogram of pairwise distances between the $k$ extracted coordinates on every benchmark: in all three datasets and every sensor budget we evaluate, no two extracted coordinates fell within a single bandwidth window of each other, which confirms that the top-$k$ values do correspond to $k$ distinct peaks rather than a cluster around the strongest one.

\paragraph{Robust fallback for higher-density regimes.} The argument above assumes that bandwidth $\sigma_x$ stays comfortably below the inter-sensor spacing. In settings where sensor density is forced higher than what the chosen bandwidth supports --- for example, packing more than $\sim\!300$ sensors onto our $128\!\times\!48$ turbulence grid --- two peaks could merge and naive top-$k$ would return multiple points around the strongest, missing some of the others. The standard remedy is a non-maximum suppression (NMS) pass, e.g.\ a $3\!\times\!3$ local-maximum filter applied to $\widehat{\mathcal{S}}_t$ before ranking, which guarantees one extracted point per local maximum. We did not require this fallback in any of our reported benchmarks, but flag it as a drop-in replacement for the readout in regimes that violate the bandwidth-vs-density assumption.

\subsection{End-to-End Training and Inference Algorithm}
\label{app:impl_algo}

For completeness we summarize the full DyPSI pipeline as a four-phase procedure (Algorithm~\ref{alg:dypsi}). Phase~1 (data preparation) generates paired field/sensor sequences by applying the model-agnostic trajectory optimizer described in \S\ref{app:data_prep} and Algorithm~\ref{alg:traj_opt} to any pre-trained reconstructor (e.g., PhySense, DiffusionPDE, or Senseiver), yielding temporally coherent and kinematically feasible sensor trajectories alongside the ground-truth physical fields. The remaining three phases constitute our core method: Functional Tucker pre-training (Phase~2), joint diffusion training (Phase~3), and conditional sampling at inference (Phase~4). Symbols not redefined here follow the main text.

\begin{algorithm}[h]
\caption{DyPSI: Functional Tucker pre-training, joint diffusion training, and conditional sampling.}
\label{alg:dypsi}
\textbf{Phase 1 (Trajectory-supervision construction.).} Construct pairs of field snapshots $\{\mathcal{P}_{1:T_n}^{(n)}\}$ and supervision sensor coordinates $\{\mathbf{s}_{1:T_n}^{(n)}\}$ from costy offline optimizer, and split into train/val/test. See details at Algorithm~\ref{alg:traj_opt}.

\medskip
\noindent\textbf{Phase 2: Functional Tucker pre-training.}
\begin{algorithmic}[1]
\State Initialize shared spatial basis MLPs $\boldsymbol{\phi}_2,\ldots,\boldsymbol{\phi}_{D+1}$, the temporal basis $\boldsymbol{\phi}_1$, and per-sample core tensors $\mathcal{W}^{\mathrm{phy}}_n,\mathcal{W}^{\mathrm{sen}}_n$.
\State For every training sample $n$ and timestep $t$, compute the SPF $\mathcal{S}^{(n)}_t$ from $\mathbf{s}^{(n)}_t$ via Eq.~\eqref{eq:method_spf}.
\State Jointly optimize $\{\boldsymbol{\phi}_k\}$ and $\{\mathcal{W}^{\mathrm{phy}}_n,\mathcal{W}^{\mathrm{sen}}_n\}$ to minimize the dual reconstruction loss
\Statex \quad $\sum_{n}\bigl(\|\mathrm{Tucker}(\mathcal{W}^{\mathrm{phy}}_n,\boldsymbol{\phi})-\mathcal{P}^{(n)}\|^{2} + \|\mathrm{Tucker}(\mathcal{W}^{\mathrm{sen}}_n,\boldsymbol{\phi})-\mathcal{S}^{(n)}\|^{2}\bigr)$
\Statex against arbitrary continuous-coordinate samples until convergence.
\State \textbf{Freeze} $\{\boldsymbol{\phi}_k\}$. Cache the resulting per-sample joint cores $\mathbf{x}_{\mathrm{joint}}^{(n)}=\mathrm{Concat}_{R_1}\!\bigl[\mathcal{W}^{\mathrm{phy}}_n,\mathcal{W}^{\mathrm{sen}}_n\bigr]$ (Eq.~\eqref{eq:method_joint_core}).
\end{algorithmic}

\medskip
\noindent\textbf{Phase 3: Joint diffusion training.}
\begin{algorithmic}[1]
\State Precompute the Mat\'ern-$3/2$ kernel matrix $\mathbf{K}_{\mathrm{time}}\!\in\!\mathbb{R}^{T_{\mathrm{seg}}\times T_{\mathrm{seg}}}$ and its Cholesky factor $\mathbf{L}$.
\For{step $=1,\ldots,N_{\mathrm{iter}}$}
  \State Sample a sample $n$ and a length-$T_{\mathrm{seg}}$ segment $i$ of $\mathbf{x}_{\mathrm{joint}}^{(n)}$ with its observed prefix $\mathcal{H}^{(i)}$ of length $K_{\mathrm{obs}}$.
  \State Encode the prefix to $\mathbf{h}_{\mathrm{hist}}^{(i)}$ with the history Transformer.
  \State Draw $\sigma\sim p(\sigma)$ and $\boldsymbol{\xi}^{(i)}\sim\mathcal{N}(\mathbf{0},\mathbf{I})$; form $\boldsymbol{\eta}^{(i)}=(\mathbf{L}\otimes \mathbf{I})\,\mathrm{vec}(\boldsymbol{\xi}^{(i)})$ (Eq.~\eqref{eq:method_corr_noise}).
  \State Update $\theta$ by stochastic gradient on $\mathcal{L}_{\mathrm{joint}}$ in Eq.~\eqref{eq:method_joint_diffusion}.
\EndFor
\end{algorithmic}

\medskip
\noindent\textbf{Phase 4: Conditional sampling and sensor extraction.}
\begin{algorithmic}[1]
\Require observed prefix $\{(\mathbf{s}_\tau,\mathbf{y}_\tau)\}_{\tau=1}^{K_{\mathrm{obs}}}$, sensor budget $k$.
\State Encode prefix $\to$ $\mathbf{h}_{\mathrm{hist}}^{(i)}$. Initialize the suffix at the highest noise level: $\mathbf{x}_R^{(i)}\sim\mathcal{N}(\mathbf{0},\sigma_R^{2}\mathbf{I})$.
\For{$r=R,\ldots,1$}
  \State Soft-anchor the prefix: $\mathbf{x}_{r,\,1:K_{\mathrm{obs}}}^{(i)}\!\leftarrow\!(1{-}\alpha_r)\mathbf{x}_{r,\,1:K_{\mathrm{obs}}}^{(i)}+\alpha_r\widetilde{\mathbf{x}}_{\mathrm{hist},r}^{(i)}$ (Eq.~\eqref{eq:method_guidance}).
  \State Reverse step: $\mathbf{x}_{r-1}^{(i)}\!\leftarrow$ second-order Heun update of $\mathbf{x}_{r}^{(i)}$ using $D_\theta(\,\cdot\,;\sigma_{r},\mathbf{h}_{\mathrm{hist}}^{(i)})$.
\EndFor
\State Decode $\mathbf{x}_0^{(i)}$ via the frozen Functional Tucker basis to recover $\widehat{\mathcal{P}}_{K_{\mathrm{obs}}+1:T_{\mathrm{seg}}}$ and $\widehat{\mathcal{S}}_{K_{\mathrm{obs}}+1:T_{\mathrm{seg}}}$.
\State \textbf{return} predicted fields $\{\widehat{\mathcal{P}}_t\}$ and sensor coordinates $\{\widehat{\mathbf{s}}_t\}$, where $\widehat{\mathbf{s}}_t$ is the top-$k$ $\arg\max$ of $\widehat{\mathcal{S}}_t$ on $\Omega$ (\S\ref{app:impl_sample}).
\end{algorithmic}
\end{algorithm}

\subsection{Per-Benchmark Hyperparameters}
\label{app:impl_hp}

Table~\ref{tab:hp} summarizes the benchmark-specific architectural and dataset hyperparameters for the three physics-sensing tasks evaluated in \S\ref{sec:experiments}. The modality channel ranks $(R_1^{\mathrm{phy}}, R_1^{\mathrm{sen}})$ control the expressiveness of the physical field and sensor position field branches within the shared Functional Tucker latent space; we allocate a higher channel rank to the physical field ($R_1^{\mathrm{phy}} > R_1^{\mathrm{sen}}$) because the physical dynamics are typically more complex than the sensor placement patterns. 
Spatial ranks $(R_2,\dots,R_{D+1})$ are chosen proportional to the spatial resolution of each benchmark. 
The SPF bandwidth is set to approximately $1.5\times$ the local grid spacing so that each Gaussian blob covers roughly a $9\times9$ neighborhood while keeping inter-sensor overlap negligible at the evaluated densities.
At inference, test trajectories longer than the training segment length are generated autoregressively in sliding windows: the final $K_{\mathrm{obs}}$ frames of each generated segment serve as the observed historical prefix for the next segment, repeating until the full test sequence is covered.

\subsection{Compute}
\label{app:impl_compute}

All experiments were run on a single NVIDIA A100-80GB. Training takes approximately 14 hours for the turbulent-flow benchmark, 22 hours for the sea-surface temperature benchmark, and 28 hours for the car-aerodynamics benchmark. Inference (per test trajectory of length $T{=}64$, sensor count $k{=}100$) takes 1.7--3.5 seconds depending on benchmark.

\section{Full Experimental Results}
\label{app:full_results}

This appendix reports the complete experimental results corresponding to the main-text comparisons. All stochastic experiments are repeated with five independent random seeds, and entries denote mean $\pm$ standard deviation. For the reconstruction benchmarks, statistical significance is assessed using paired $t$-tests between DyPSI and the strongest baseline in each setting on identical test cases. The main-text tables report mean values for compactness, while the following tables provide the complete variability and significance results.

Table~\ref{tab:app_main_2d_full} confirms that DyPSI achieves the lowest mean error in every turbulent flow and sea surface temperature configuration. The standard deviations are small relative to the corresponding improvements over PhySense-opt. All turbulent flow comparisons are significant at $p<0.001$, and the sea surface temperature improvements remain significant with $p$-values between 0.004 and 0.010.

On car aerodynamics, Table~\ref{tab:app_car_full} shows consistent improvements of 17.47\% to 19.43\% over PhySense-opt. Every comparison is significant at $p<0.001$. The DyPSI standard deviations range from 0.0011 to 0.0014, indicating stable performance across all sensor budgets.

Table~\ref{tab:app_kinematic_tradeoff_full} provides the complete variability results for the accuracy and kinematic feasibility study. The Per-frame Oracle achieves the lowest reconstruction error but violates the velocity constraint in 75\% to 92\% of transitions and produces substantially larger smoothness costs. DyPSI reduces the violation rate to 4\% to 9\% while improving over static PhySense-opt and remaining close to the ground-truth-informed Sequence Teacher. Its runtime of 1.9 to 2.8 seconds per sequence is approximately three orders of magnitude lower than that of the optimization-based alternatives.

Table~\ref{tab:app_placement_generalization_full} shows that DyPSI-opt attains the lowest mean error across all teacher architectures and sensor budgets. Eleven of the twelve comparisons against PhySense-opt are significant at $p\leq0.007$. The Senseiver comparison with 30 sensors is not statistically significant, with $p=0.400$, and the corresponding mean reduction is substantially smaller than in the other settings. Overall, the results support generalization across reconstruction architectures while also identifying the configuration where the observed improvement is not statistically resolved.

\begin{sidewaystable}[p]
\centering
\scriptsize
\setlength{\tabcolsep}{2.2pt}
\caption{Full performance comparison on the turbulent flow and sea surface temperature benchmarks. Relative $L_2$ error is reported as mean $\pm$ standard deviation over five independent runs. Improvement quantifies the relative error reduction with respect to PhySense-opt. The $p$-values are obtained from paired $t$-tests comparing DyPSI with PhySense-opt on identical test cases.}
\label{tab:app_main_2d_full}
\resizebox{0.98\textheight}{!}{%
\begin{tabular}{ll|cccc|cccc}
\toprule
 &  & \multicolumn{4}{c|}{Turbulent Flow} & \multicolumn{4}{c}{Sea Temperature} \\
Models & Placement & \#200 & \#100 & \#50 & \#30 & \#100 & \#50 & \#25 & \#15 \\
\midrule
SSPOR & SSPOR & 0.3018 $\pm$ 0.0057 & 0.4429 $\pm$ 0.0071 & 0.5921 $\pm$ 0.0105 & 0.6637 $\pm$ 0.0118 & 0.0756 $\pm$ 0.0026 & 0.0752 $\pm$ 0.0029 & 0.0789 $\pm$ 0.0026 & 0.0719 $\pm$ 0.0034 \\
VoronoiCNN & Random & 0.3588 $\pm$ 0.0041 & 0.4870 $\pm$ 0.0069 & 0.6919 $\pm$ 0.0108 & 0.7992 $\pm$ 0.0130 & 0.1496 $\pm$ 0.0041 & 0.1537 $\pm$ 0.0048 & 0.2433 $\pm$ 0.0063 & 0.2817 $\pm$ 0.0071 \\
Senseiver & Random & 0.1842 $\pm$ 0.0032 & 0.2316 $\pm$ 0.0049 & 0.3740 $\pm$ 0.0069 & 0.5746 $\pm$ 0.0086 & 0.0715 $\pm$ 0.0020 & 0.0732 $\pm$ 0.0021 & 0.0769 $\pm$ 0.0027 & 0.0784 $\pm$ 0.0028 \\
S$^3$GM & Random & 0.1856 $\pm$ 0.0031 & 0.2016 $\pm$ 0.0035 & 0.2681 $\pm$ 0.0042 & 0.6421 $\pm$ 0.0098 & 0.1115 $\pm$ 0.0034 & 0.1137 $\pm$ 0.0038 & 0.1581 $\pm$ 0.0045 & 0.1795 $\pm$ 0.0050 \\
DiffusionPDE & Random & 0.1993 $\pm$ 0.0035 & 0.3312 $\pm$ 0.0061 & 0.6604 $\pm$ 0.0103 & 0.9163 $\pm$ 0.0125 & 0.0664 $\pm$ 0.0025 & 0.0775 $\pm$ 0.0031 & 0.0859 $\pm$ 0.0029 & 0.1139 $\pm$ 0.0032 \\
PhySense & Random & 0.1233 $\pm$ 0.0022 & 0.1527 $\pm$ 0.0028 & 0.2586 $\pm$ 0.0047 & 0.5176 $\pm$ 0.0083 & 0.0439 $\pm$ 0.0016 & 0.0452 $\pm$ 0.0019 & 0.0477 $\pm$ 0.0023 & 0.0520 $\pm$ 0.0014 \\
PhySense-opt & Static Opt. & 0.1106 $\pm$ 0.0018 & 0.1257 $\pm$ 0.0020 & 0.1558 $\pm$ 0.0027 & 0.2157 $\pm$ 0.0045 & 0.0426 $\pm$ 0.0015 & 0.0430 $\pm$ 0.0016 & 0.0437 $\pm$ 0.0021 & 0.0439 $\pm$ 0.0018 \\
\textbf{DyPSI} & Dynamic Gen. & \textbf{0.0986 $\pm$ 0.0013} & \textbf{0.1101 $\pm$ 0.0017} & \textbf{0.1352 $\pm$ 0.0024} & \textbf{0.1826 $\pm$ 0.0041} & \textbf{0.0387 $\pm$ 0.0012} & \textbf{0.0390 $\pm$ 0.0014} & \textbf{0.0394 $\pm$ 0.0019} & \textbf{0.0402 $\pm$ 0.0020} \\
\midrule
\multicolumn{2}{l|}{Improvement} & 10.87\% & 12.41\% & 13.19\% & 15.35\% & 9.15\% & 9.30\% & 9.84\% & 8.43\% \\
\multicolumn{2}{l|}{$p$-value} & $<0.001$ & $<0.001$ & $<0.001$ & $<0.001$ & 0.004 & 0.004 & 0.005 & 0.010 \\
\bottomrule
\end{tabular}%
}

\vspace{2em}

\caption{Full accuracy and kinematic feasibility comparison under representative sparse sensor budgets. Entries denote mean $\pm$ standard deviation over five independent runs. Viol. denotes the velocity violation rate, Smooth. denotes $\sum_t\|\mathbf{s}_{t+1}-2\mathbf{s}_t+\mathbf{s}_{t-1}\|^2$, and runtime is measured in seconds per 64-frame sequence. The Per-frame Oracle and Sequence Teacher access future ground truth and perform test-time optimization, whereas DyPSI uses neither.}
\label{tab:app_kinematic_tradeoff_full}
\resizebox{0.98\textheight}{!}{%
\begin{tabular}{lcc|cccc|cccc|cccc}
\toprule
 &  &  & \multicolumn{4}{c|}{Turbulent Flow, 50 sensors} & \multicolumn{4}{c|}{Sea Surface Temperature, 25 sensors} & \multicolumn{4}{c}{Car Aerodynamics, 30 sensors} \\
Strategy & Future GT & Test-time Opt. & Rel. $L_2$ $\downarrow$ & Viol. $\downarrow$ & Smooth. $\downarrow$ & Runtime $\downarrow$ & Rel. $L_2$ $\downarrow$ & Viol. $\downarrow$ & Smooth. $\downarrow$ & Runtime $\downarrow$ & Rel. $L_2$ $\downarrow$ & Viol. $\downarrow$ & Smooth. $\downarrow$ & Runtime $\downarrow$ \\
\midrule
Static PhySense-opt & $\times$ & $\times$ & 0.1558 $\pm$ 0.0027 & 0 & 0 & $-$ & 0.0437 $\pm$ 0.0018 & 0 & 0 & $-$ & 0.0372 $\pm$ 0.0015 & 0 & 0 & $-$ \\
Per-frame Oracle & $\checkmark$ & $\checkmark$ & 0.1007 $\pm$ 0.0012 & 89\% $\pm$ 5\% & 142.5 $\pm$ 13.8 & 887 $\pm$ 51 & 0.0259 $\pm$ 0.0015 & 75\% $\pm$ 3\% & 165.3 $\pm$ 14.2 & 628 $\pm$ 39 & 0.0201 $\pm$ 0.0017 & 92\% $\pm$ 3\% & 88.6 $\pm$ 9.5 & 1026 $\pm$ 103 \\
Sequence Teacher & $\checkmark$ & $\checkmark$ & 0.1206 $\pm$ 0.0009 & 5\% $\pm$ 1\% & 3.5 $\pm$ 0.9 & 1683 $\pm$ 143 & 0.0355 $\pm$ 0.0011 & 2\% $\pm$ 1\% & 4.5 $\pm$ 1.1 & 1429 $\pm$ 107 & 0.0251 $\pm$ 0.0012 & 6\% $\pm$ 2\% & 2.6 $\pm$ 0.7 & 1952 $\pm$ 165 \\
\textbf{DyPSI} & $\times$ & $\times$ & 0.1352 $\pm$ 0.0024 & 8\% $\pm$ 2\% & 4.2 $\pm$ 1.0 & \textbf{1.9 $\pm$ 0.2} & 0.0394 $\pm$ 0.0019 & 4\% $\pm$ 1\% & 9.2 $\pm$ 1.9 & \textbf{2.1 $\pm$ 0.2} & 0.0307 $\pm$ 0.0014 & 9\% $\pm$ 2\% & 3.1 $\pm$ 0.8 & \textbf{2.8 $\pm$ 0.3} \\
\bottomrule
\end{tabular}%
}
\end{sidewaystable}

\begin{sidewaystable}[p]
\centering
\scriptsize
\setlength{\tabcolsep}{2.0pt}
\caption{Full constructor/backbone generalization results on the turbulent flow benchmark. Relative $L_2$ error is reported as mean $\pm$ standard deviation over five independent runs. For non-DyPSI rows, column groups denote the reconstruction backbone used for evaluation. For DyPSI-opt, they denote the teacher used to construct offline trajectory supervision for a separately trained DyPSI model. The $p$-values compare DyPSI-opt with PhySense-opt in each setting.}
\label{tab:app_placement_generalization_full}
\resizebox{0.98\textheight}{!}{%
\begin{tabular}{l|cccc|cccc|cccc}
\toprule
 & \multicolumn{4}{c|}{Senseiver} & \multicolumn{4}{c|}{PhySense} & \multicolumn{4}{c}{DiffusionPDE} \\
Placement Strategy & \#200 & \#100 & \#50 & \#30 & \#200 & \#100 & \#50 & \#30 & \#200 & \#100 & \#50 & \#30 \\
\midrule
Random Sample & 0.1842 $\pm$ 0.0037 & 0.2316 $\pm$ 0.0043 & 0.3740 $\pm$ 0.0053 & 0.5746 $\pm$ 0.0090 & 0.1233 $\pm$ 0.0021 & 0.1527 $\pm$ 0.0031 & 0.2586 $\pm$ 0.0040 & 0.5176 $\pm$ 0.0073 & 0.1993 $\pm$ 0.0046 & 0.3312 $\pm$ 0.0052 & 0.6604 $\pm$ 0.0105 & 0.9163 $\pm$ 0.0143 \\
Grid Sample & 0.1615 $\pm$ 0.0030 & 0.2273 $\pm$ 0.0041 & 0.3845 $\pm$ 0.0062 & 0.5869 $\pm$ 0.0092 & 0.1235 $\pm$ 0.0029 & 0.1521 $\pm$ 0.0036 & 0.2436 $\pm$ 0.0047 & 0.4930 $\pm$ 0.0068 & 0.1932 $\pm$ 0.0031 & 0.3319 $\pm$ 0.0071 & 0.6703 $\pm$ 0.0093 & 0.9079 $\pm$ 0.0115 \\
Min-Max Sample & 0.1549 $\pm$ 0.0028 & 0.2091 $\pm$ 0.0038 & 0.3795 $\pm$ 0.0067 & 0.5747 $\pm$ 0.0095 & 0.1196 $\pm$ 0.0024 & 0.1533 $\pm$ 0.0020 & 0.3157 $\pm$ 0.0052 & 0.5022 $\pm$ 0.0041 & 0.1889 $\pm$ 0.0037 & 0.3307 $\pm$ 0.0067 & 0.6574 $\pm$ 0.0117 & 0.9010 $\pm$ 0.0132 \\
PhySense-opt & 0.1388 $\pm$ 0.0025 & 0.1750 $\pm$ 0.0032 & 0.3025 $\pm$ 0.0054 & 0.4570 $\pm$ 0.0079 & 0.1106 $\pm$ 0.0014 & 0.1257 $\pm$ 0.0021 & 0.1558 $\pm$ 0.0031 & 0.2157 $\pm$ 0.0037 & 0.1672 $\pm$ 0.0025 & 0.2841 $\pm$ 0.0052 & 0.5319 $\pm$ 0.0085 & 0.7904 $\pm$ 0.0120 \\
\textbf{DyPSI-opt} & \textbf{0.1259 $\pm$ 0.0022} & \textbf{0.1670 $\pm$ 0.0030} & \textbf{0.2863 $\pm$ 0.0048} & \textbf{0.4527 $\pm$ 0.0072} & \textbf{0.0986 $\pm$ 0.0018} & \textbf{0.1101 $\pm$ 0.0020} & \textbf{0.1352 $\pm$ 0.0028} & \textbf{0.1826 $\pm$ 0.0041} & \textbf{0.1569 $\pm$ 0.0029} & \textbf{0.2640 $\pm$ 0.0048} & \textbf{0.5028 $\pm$ 0.0075} & \textbf{0.7116 $\pm$ 0.0110} \\
\midrule
$p$-value & $<0.001$ & 0.007 & 0.003 & 0.400 & $<0.001$ & $<0.001$ & $<0.001$ & $<0.001$ & 0.003 & 0.001 & 0.002 & $<0.001$ \\
\bottomrule
\end{tabular}%
}

\vspace{2em}

\caption{Full gain decomposition on all reconstruction benchmarks. Errors are reported as mean $\pm$ standard deviation over five independent runs. Gain Retention is computed using Eq.~\eqref{eq:gain_retention}. TF and SST denote turbulent flow and sea surface temperature, respectively.}
\label{tab:app_gain_decomposition}
\resizebox{0.98\textheight}{!}{%
\begin{tabular}{l|cccc|cccc|cccc}
\toprule
 & \multicolumn{4}{c|}{Turbulent Flow} & \multicolumn{4}{c|}{Sea Surface Temperature} & \multicolumn{4}{c}{Car Aerodynamics} \\
Strategy & \#200 & \#100 & \#50 & \#30 & \#100 & \#50 & \#25 & \#15 & \#100 & \#50 & \#30 & \#15 \\
\midrule
Static PhySense-opt & 0.1106 $\pm$ 0.0018 & 0.1257 $\pm$ 0.0020 & 0.1558 $\pm$ 0.0027 & 0.2157 $\pm$ 0.0045 & 0.0426 $\pm$ 0.0015 & 0.0430 $\pm$ 0.0016 & 0.0437 $\pm$ 0.0021 & 0.0439 $\pm$ 0.0018 & 0.0370 $\pm$ 0.0012 & 0.0370 $\pm$ 0.0016 & 0.0372 $\pm$ 0.0015 & 0.0386 $\pm$ 0.0021 \\
Sequence Teacher & 0.0915 $\pm$ 0.0008 & 0.1025 $\pm$ 0.0015 & 0.1206 $\pm$ 0.0009 & 0.1695 $\pm$ 0.0019 & 0.0318 $\pm$ 0.0018 & 0.0347 $\pm$ 0.0011 & 0.0355 $\pm$ 0.0011 & 0.0371 $\pm$ 0.0016 & 0.0230 $\pm$ 0.0008 & 0.0247 $\pm$ 0.0017 & 0.0251 $\pm$ 0.0012 & 0.0273 $\pm$ 0.0019 \\
\textbf{DyPSI} & 0.0986 $\pm$ 0.0013 & 0.1101 $\pm$ 0.0017 & 0.1352 $\pm$ 0.0024 & 0.1826 $\pm$ 0.0041 & 0.0387 $\pm$ 0.0012 & 0.0390 $\pm$ 0.0014 & 0.0394 $\pm$ 0.0019 & 0.0402 $\pm$ 0.0020 & 0.0301 $\pm$ 0.0013 & 0.0302 $\pm$ 0.0011 & 0.0307 $\pm$ 0.0014 & 0.0311 $\pm$ 0.0014 \\
\midrule
\textbf{Gain Retention} & \textbf{0.63} & \textbf{0.67} & \textbf{0.59} & \textbf{0.72} & \textbf{0.36} & \textbf{0.48} & \textbf{0.52} & \textbf{0.54} & \textbf{0.49} & \textbf{0.55} & \textbf{0.54} & \textbf{0.66} \\
\bottomrule
\end{tabular}%
}
\end{sidewaystable}

\clearpage

\begin{table}[t]
\centering
\footnotesize
\setlength{\tabcolsep}{3.0pt}
\caption{Full results on the car aerodynamics benchmark. Relative $L_2$ error is reported as mean $\pm$ standard deviation over five independent runs. The $p$-values are obtained from paired $t$-tests comparing DyPSI with PhySense-opt.}
\label{tab:app_car_full}
\resizebox{\textwidth}{!}{%
\begin{tabular}{l|ccccc}
\toprule
Models & \#200 & \#100 & \#50 & \#30 & \#15 \\
\midrule
Senseiver & 0.1009 $\pm$ 0.0030 & 0.1012 $\pm$ 0.0031 & 0.1018 $\pm$ 0.0037 & 0.1022 $\pm$ 0.0030 & 0.1053 $\pm$ 0.0032 \\
DiffusionPDE & 0.0967 $\pm$ 0.0028 & 0.0966 $\pm$ 0.0028 & 0.0987 $\pm$ 0.0039 & 0.1055 $\pm$ 0.0031 & 0.2095 $\pm$ 0.0060 \\
PhySense & 0.0375 $\pm$ 0.0013 & 0.0382 $\pm$ 0.0013 & 0.0395 $\pm$ 0.0017 & 0.0416 $\pm$ 0.0015 & 0.0465 $\pm$ 0.0023 \\
PhySense-opt & 0.0369 $\pm$ 0.0012 & 0.0370 $\pm$ 0.0012 & 0.0370 $\pm$ 0.0016 & 0.0372 $\pm$ 0.0015 & 0.0386 $\pm$ 0.0021 \\
\textbf{DyPSI} & \textbf{0.0301 $\pm$ 0.0013} & \textbf{0.0301 $\pm$ 0.0013} & \textbf{0.0302 $\pm$ 0.0011} & \textbf{0.0307 $\pm$ 0.0014} & \textbf{0.0311 $\pm$ 0.0014} \\
\midrule
Improvement & 18.43\% & 18.65\% & 18.39\% & 17.47\% & 19.43\% \\
$p$-value & $<0.001$ & $<0.001$ & $<0.001$ & $<0.001$ & $<0.001$ \\
\bottomrule
\end{tabular}%
}
\end{table}

\subsection{Gain Decomposition Relative to the Offline Teacher}
\label{app:gain_decomposition}

We further quantify how much of the offline teacher's improvement over static placement is retained by the learned DyPSI generator. Static PhySense-opt, the Sequence Teacher, and DyPSI are evaluated on identical test sequences with the same frozen reconstructor. The Sequence Teacher has access to future ground-truth fields and performs per-sequence optimization, whereas DyPSI uses only historical observations and performs no test-time optimization. Let $E_{\mathrm{static}}$, $E_{\mathrm{teacher}}$, and $E_{\mathrm{DyPSI}}$ denote their respective relative $L_2$ errors. We define
\begin{equation}
\mathrm{GainRetention}
=\frac{E_{\mathrm{static}}-E_{\mathrm{DyPSI}}}
{E_{\mathrm{static}}-E_{\mathrm{teacher}}}.
\label{eq:gain_retention}
\end{equation}
This metric measures the fraction of the teacher's total error reduction over static placement that remains after amortizing the offline optimizer into the history-conditioned generator. The difference between static placement and the teacher reflects the value of dynamic sequence-level sensing with oracle information, while the difference between the teacher and DyPSI reflects the amortization gap introduced by removing future ground-truth access and online optimization.

Table~\ref{tab:app_gain_decomposition} shows that DyPSI retains 36\% to 72\% of the teacher's gain across the twelve configurations, with an average retention of approximately 56\%. The teacher consistently achieves lower error, as expected from its access to future ground truth and its substantially larger optimization budget. Nevertheless, DyPSI recovers roughly half of the available dynamic improvement without either advantage. This decomposition demonstrates that the gains arise from both sequence-level adaptive sensing and the successful transfer of the teacher's strategy to the learned generator, while the remaining error difference quantifies the cost of amortization.

\section{Construct Trajectory Supervision Data}
\label{app:data_prep}

\paragraph{Motivation.}
Public physics-sensing datasets---including the GLORYS12 reanalysis, OpenFOAM car-aerodynamics simulations, and the turbulent channel-flow snapshots used in \S\ref{sec:experiments}---provide dense spatiotemporal fields but do not contain paired sensor trajectories. To train DyPSI's joint generative model, we require paired sequences $\{(\mathcal{P}_{1:T},\mathbf{s}_{1:T})\}$ in which the sensing configurations are not only informative but also temporally coherent and kinematically feasible for mobile platforms. We therefore construct high-quality dynamic trajectories via an offline sequence-level optimization that can be instantiated on top of \emph{any} differentiable reconstructor.

\paragraph{Model-agnostic trajectory optimization.}
Given a pre-trained reconstructor $\mathcal{M}_{\theta}$ (which may be PhySense, DiffusionPDE, Senseiver, or any other end-to-end differentiable model), a ground-truth field sequence $\{\mathcal{P}_t^{\text{gt}}\}_{t=1}^{T}$, and an initial static placement $\mathbf{s}_1^{\star}$ obtained from any frame-wise optimizer (e.g., the PGD stage of PhySense-opt), we seek a dynamic trajectory $\mathbf{S}_{1:T}=\{\mathbf{s}_t\}_{t=1}^{T}$ that minimizes the composite objective
\begin{equation}
\mathcal{L}_{\text{total}}(\mathbf{S}_{1:T}) \;=\; \underbrace{\mathcal{L}_{\text{recon}}}_{\text{reconstruction fidelity}} \;+\; \lambda\,\underbrace{\mathcal{L}_{\text{GP}}}_{\text{temporal coherence}} \;+\; \gamma\,\underbrace{\mathcal{L}_{\text{dyn}}}_{\text{kinematic feasibility}} .
\label{eq:data_total_loss}
\end{equation}

The reconstruction term measures the cumulative sparse-to-dense error over the sequence:
\begin{equation}
\mathcal{L}_{\text{recon}} \;=\; \sum_{t=1}^{T}\bigl\| \mathcal{M}_{\theta}\bigl(\mathcal{P}_t^{\text{gt}}(\mathbf{s}_t);\,\mathbf{s}_t\bigr) \;-\; \mathcal{P}_t^{\text{gt}} \bigr\|_2^{2},
\label{eq:data_recon_loss}
\end{equation}
where $\mathcal{P}_t^{\text{gt}}(\mathbf{s}_t)$ denotes the sparse point observations sampled at coordinates $\mathbf{s}_t$. Because $\mathcal{M}_{\theta}$ is differentiable, gradients flow end-to-end from the reconstruction loss back to the sensor coordinates; when the reconstructor operates on the Sensor Position Field surrogate $\mathcal{S}_t$ (\S3.2), the map $\mathbf{s}_t\!\to\!\mathcal{S}_t$ in Eq.~(7) supplies a fully differentiable bridge that avoids discrete argmax operations during optimization.

The temporal term is the GP log-prior adopted in DyPSI's diffusion training (Eq.~(10)):
\begin{equation}
\mathcal{L}_{\text{GP}} \;=\; \frac{1}{2}\sum_{i=1}^{k}\bigl(\mathbf{s}_{1:T}^{(i)}-\mathbf{m}\bigr)^{\!\top}\mathbf{K}^{-1}\bigl(\mathbf{s}_{1:T}^{(i)}-\mathbf{m}\bigr),
\label{eq:data_gp_loss}
\end{equation}
where $\mathbf{s}_{1:T}^{(i)}\in\mathbb{R}^{T\times 2}$ is the trajectory of the $i$-th sensor, $\mathbf{m}$ is a reference path (we use the initial static placement as mean), and $\mathbf{K}\in\mathbb{R}^{T\times T}$ is the Mat\'ern-$3/2$ Gram matrix with length scale $\ell=4$ frames and signal variance $\sigma_{\mathrm{GP}}^{2}=1$, identical to the kernel used in \S3.3. Crucially, this is the \emph{same} kernel family analysed in Proposition~2 (\S3.4): its quadratic form furnishes a principled upper bound on the discrete frame-to-frame variation of the trajectory, thereby guaranteeing kinematic feasibility without any ad hoc smoothing penalty.

The dynamics term enforces hard platform constraints via hinge losses on velocity and acceleration:
\begin{equation}
\begin{aligned}
\mathcal{L}_{\text{dyn}} \;=\; \sum_{t=1}^{T-1}\!\mathrm{ReLU}\!\Bigl(\|\mathbf{s}_{t+1}-\mathbf{s}_{t}\|-v_{\max}\Delta t\Bigr)^{2}
\;+\; \sum_{t=2}^{T-1}\!\mathrm{ReLU}\!\Bigl(\|\mathbf{s}_{t+1}-2\mathbf{s}_{t}+\mathbf{s}_{t-1}\|-a_{\max}\Delta t^{2}\Bigr)^{2},
\end{aligned}
\label{eq:data_dyn_loss}
\end{equation}
where $v_{\max}$ and $a_{\max}$ are platform-specific speed and acceleration limits. The full procedure is summarized in Algorithm~\ref{alg:traj_opt}.

\begin{algorithm}[h]
\caption{Sequence-Level Sensor-Trajectory Optimization (Offline Data Generator)}
\label{alg:traj_opt}
\begin{algorithmic}[1]
\Require Pre-trained reconstructor $\mathcal{M}_{\theta}$; ground-truth fields $\{\mathcal{P}_t^{\text{gt}}\}_{t=1}^{T}$; initial static placement $\mathbf{s}_1^{\star}$ from PhySense-opt (or any frame-wise optimizer); sensor budget $k$; Mat\'ern-$3/2$ kernel matrix $\mathbf{K}$ with Cholesky factor $\mathbf{L}$; hyperparameters $\lambda,\gamma,v_{\max},a_{\max}$; step size $\eta$; iterations $N_{\text{iter}}$.
\Ensure Optimized dynamic trajectory $\mathbf{S}_{1:T}=\{\mathbf{s}_t\}_{t=1}^{T}$.
\State Initialize $\mathbf{s}_1\leftarrow\mathbf{s}_1^{\star}$; for $t=2,\dots,T$ initialize $\mathbf{s}_t$ by optical-flow advection from $\mathbf{s}_{t-1}$ or by sampling from the GP prior.
\For{iteration $=1,\dots,N_{\text{iter}}$}
    \State $\mathcal{L}_{\text{recon}}\leftarrow\sum_{t=1}^{T}\bigl\|\mathcal{M}_{\theta}\bigl(\mathcal{P}_t^{\text{gt}}(\mathbf{s}_t);\mathbf{s}_t\bigr)-\mathcal{P}_t^{\text{gt}}\bigr\|_2^{2}$
    \State $\mathcal{L}_{\text{GP}}\leftarrow\frac{1}{2}\sum_{i=1}^{k}\bigl\|\mathbf{L}^{-1}\bigl(\mathbf{s}_{1:T}^{(i)}-\mathbf{m}\bigr)\bigr\|_2^{2}$
    \State $\mathcal{L}_{\text{dyn}}\leftarrow\text{velocity\&acceleration\ hinge\ losses\ via\ Eq.~\eqref{eq:data_dyn_loss}}$
    \State $\mathcal{L}_{\text{total}}\leftarrow\mathcal{L}_{\text{recon}}+\lambda\,\mathcal{L}_{\text{GP}}+\gamma\,\mathcal{L}_{\text{dyn}}$
    \State $\mathbf{S}_{1:T}\leftarrow\mathbf{S}_{1:T}-\eta\,\nabla_{\mathbf{S}}\mathcal{L}_{\text{total}}$
    \State Project each $\mathbf{s}_t$ onto the feasible domain $\Omega$ (e.g., land masking for sea temperature, mesh surface for car aerodynamics).
\EndFor
\State \Return $\mathbf{S}_{1:T}$
\end{algorithmic}
\end{algorithm}

\paragraph{Remarks on generality.}
The pipeline above is deliberately model-agnostic: it only requires that $\mathcal{M}_{\theta}$ admit gradients with respect to the sensing coordinates (or their SPF surrogate). Consequently, the same offline generator can be paired with PhySense, DiffusionPDE, Senseiver, VoronoiCNN, or any future reconstructor without architectural modification. The initial static placement $\mathbf{s}_1^{\star}$ may likewise be produced by any frame-wise optimizer---PhySense's PGD stage, Diff-SPORT's SHAP-guided MAP-GA, or even a simple greedy A-optimal design---because the sequence-level GP prior and dynamics constraints are applied \emph{after} initialization and are independent of the static optimizer's internals.

\paragraph{Validation.}
Table~\ref{tab:data_prep} compares the reconstruction accuracy of a fixed pre-trained PhySense reconstructor under two sensing protocols on the training corpus: (i) static placements optimized by PhySense-opt's PGD stage across all frames, and (ii) dynamically optimized trajectories produced by Algorithm~\ref{alg:traj_opt}. The dynamic trajectories consistently outperform their static counterparts, with relative error reductions of approximately $20\%$ on the two grid benchmarks and $28.92\%$ on the unstructured car mesh, confirming that temporal coherence and kinematic feasibility are not merely physical desiderata but also measurably improve the information content of the sparse observations used for training.

\begin{table}[h]
\centering
\footnotesize
\caption{Training-set reconstruction error ($\downarrow$) of a fixed PhySense reconstructor under static placements versus dynamically optimized trajectories (Algorithm~\ref{alg:traj_opt}). The relative improvement quantifies the gain attained by sequence-level trajectory optimization in the data-preparation stage.}
\label{tab:data_prep}
\begin{tabular}{l|c|ccc}
\toprule
Benchmark & \#Sensors & Static (PhySense-opt) & Dynamic (Ours) & Relative Improvement \\
\midrule
Turbulent Flow & 50 & 0.1558 & 0.1206 & 22.59\% \\
Sea Temperature & 50 & 0.0430 & 0.0345 & 19.77\% \\
Car Aerodynamics & 50 & 0.0370 & 0.0263 & 28.92\% \\
\bottomrule
\end{tabular}
\end{table}

\begin{figure}[h]
\centering
\includegraphics[width=\textwidth]{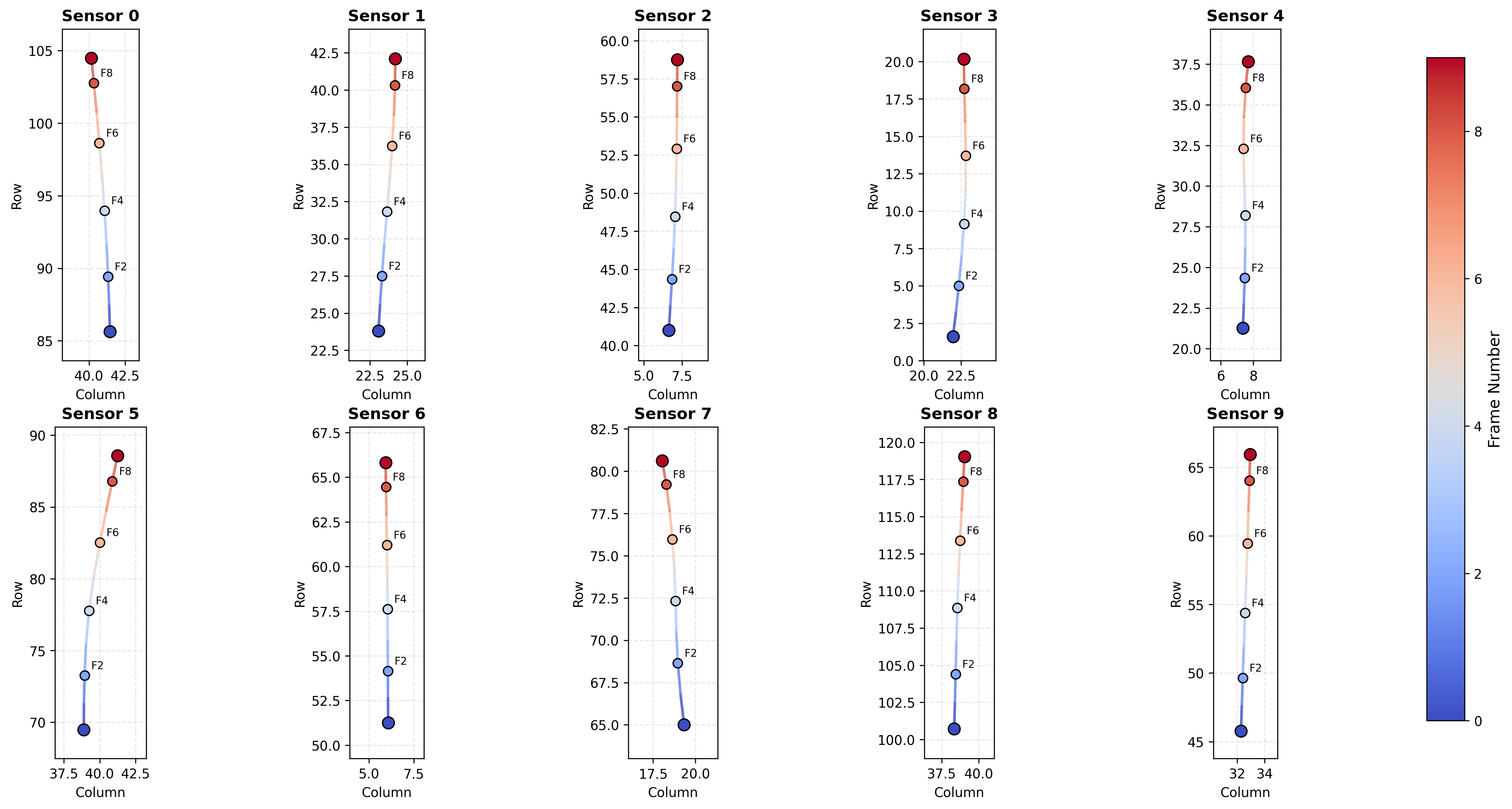}
\caption{Optimized sensor trajectories on the turbulent-flow benchmark (training-set visualization). Each panel tracks one mobile sensor across frames $0$--$9$; color encodes temporal progression from early (blue) to late (red). The smooth vertical migration demonstrates the effect of the Mat\'ern-$3/2$ GP prior and the kinematic constraints in Eq.~\eqref{eq:data_dyn_loss}.}
\label{fig:sensor_traj}
\end{figure}

Figure~\ref{fig:sensor_traj} visualizes representative trajectories generated by Algorithm~\ref{alg:traj_opt} on the turbulent-flow benchmark. Each panel tracks one mobile sensor across frames $0$--$9$; the color gradient from blue to red encodes temporal progression. The paths exhibit smooth, physically feasible vertical migration that tracks coherent flow structures, in contrast to the discontinuous ``teleporting'' behavior produced by independent per-frame optimization.

\section{Broader Impacts}
\label{app:broimp}

DyPSI advances adaptive physical sensing by enabling temporally coherent, kinematically feasible sensor trajectories that track evolving phenomena, with direct benefits for climate monitoring, oceanographic surveys, and industrial diagnostics. By maximizing information extraction from sparse mobile sensor budgets, our framework can reduce the operational costs and environmental footprint of autonomous sensing campaigns conducted by underwater vehicles, drifting buoys, or aerial platforms. We encourage the responsible deployment of such adaptive sensing systems to enhance scientific understanding and public safety, while remaining mindful of potential dual-use concerns in unauthorized tracking or surveillance scenarios.
\newpage
\section*{NeurIPS Paper Checklist}
\begin{ack}
The checklist is designed to encourage best practices for responsible machine learning research, addressing issues of reproducibility, transparency, research ethics, and societal impact. Do not remove the checklist: {\bf The papers not including the checklist will be desk rejected.} The checklist should follow the references and follow the (optional) supplemental material.  The checklist does NOT count towards the page
limit. 

Please read the checklist guidelines carefully for information on how to answer these questions. For each question in the checklist:
\begin{itemize}
    \item You should answer \answerYes{}, \answerNo{}, or \answerNA{}.
    \item \answerNA{} means either that the question is Not Applicable for that particular paper or the relevant information is Not Available.
    \item Please provide a short (1--2 sentence) justification right after your answer (even for \answerNA). 
\end{itemize}

{\bf The checklist answers are an integral part of your paper submission.} They are visible to the reviewers, area chairs, senior area chairs, and ethics reviewers. You will also be asked to include it (after eventual revisions) with the final version of your paper, and its final version will be published with the paper.

The reviewers of your paper will be asked to use the checklist as one of the factors in their evaluation. While \answerYes{} is generally preferable to \answerNo{}, it is perfectly acceptable to answer \answerNo{} provided a proper justification is given (e.g., error bars are not reported because it would be too computationally expensive'' or ``we were unable to find the license for the dataset we used''). In general, answering \answerNo{} or \answerNA{} is not grounds for rejection. While the questions are phrased in a binary way, we acknowledge that the true answer is often more nuanced, so please just use your best judgment and write a justification to elaborate. All supporting evidence can appear either in the main paper or the supplemental material, provided in appendix. If you answer \answerYes{} to a question, in the justification please point to the section(s) where related material for the question can be found.

IMPORTANT, please:
\begin{itemize}
    \item {\bf Delete this instruction block, but keep the section heading ``NeurIPS Paper Checklist"},
    \item  {\bf Keep the checklist subsection headings, questions/answers and guidelines below.}
    \item {\bf Do not modify the questions and only use the provided macros for your answers}.
\end{itemize} 
 \end{ack}


\begin{enumerate}

\item {\bf Claims}
    \item[] Question: Do the main claims made in the abstract and introduction accurately reflect the paper's contributions and scope?
    \item[] Answer: \answerYes{} 
    \item[] Justification: The abstract and introduction state the joint field--trajectory formulation, the DyPSI method, and the theoretical and empirical contributions, with evaluation on three benchmarks. The experimental claims concern improvements over random and static optimized placements and the accuracy--efficiency trade-off relative to ground-truth-informed per-frame and sequence-level oracle optimization, consistent with \S\ref{subsec:theory_motivation} and \S\ref{sec:experiments}.

\item {\bf Limitations}
    \item[] Question: Does the paper discuss the limitations of the work performed by the authors?
    \item[] Answer: \answerYes{} 
    \item[] Justification: Section 6 discusses teacher dependence and the amortization gap, the training-support assumption and out-of-distribution deployment, history-conditioned open-loop evaluation, the restriction to selected kinematic and domain constraints rather than full platform dynamics, SPF peak collisions at high sensor densities, and the point-wise observation assumption.

\item {\bf Theory assumptions and proofs}
    \item[] Question: For each theoretical result, does the paper provide the full set of assumptions and a complete (and correct) proof?
    \item[] Answer: \answerYes{} 
    \item[] Justification: All theoretical results are numbered and cross-referenced. Propositions 1–2 in the main text (\S\ref{subsec:theory_motivation}) are proved formally in Appendix ~\ref{app:theory} as Propositions A.6 and A.8. Assumptions A.1–A.3 (conditional Gaussian posterior, uniform spectral bound, admissible EDM weighting) are stated explicitly in Appendix ~\ref{app:theory}, and detailed proof sketches using spectral decomposition and generalized Rayleigh quotients are provided.

    \item {\bf Experimental result reproducibility}
    \item[] Question: Does the paper fully disclose all the information needed to reproduce the main experimental results of the paper to the extent that it affects the main claims and/or conclusions of the paper (regardless of whether the code and data are provided or not)?
    \item[] Answer: \answerYes{} 
    \item[] Justification: \S\ref{sec:experiments} describes the benchmarks, baselines, and evaluation protocol. Appendix ~\ref{app:impl} provides the metric (~\ref{app:metrics}), architectures (~\ref{app:impl_arch}), optimization and diffusion settings (~\ref{app:impl_train}), sampling procedure (~\ref{app:impl_sample}), end-to-end algorithm (~\ref{app:impl_algo}), per-benchmark hyperparameters in Table 8 (~\ref{app:impl_hp}), and compute resources (~\ref{app:impl_compute}); the public repository provides configuration files and reproduction scripts.

\item {\bf Open access to data and code}
    \item[] Question: Does the paper provide open access to the data and code, with sufficient instructions to faithfully reproduce the main experimental results, as described in supplemental material?
    \item[] Answer: \answerYes{} 
    \item[] Justification: The public code repository is available at 
    \url{https://github.com/mengqi132/DyPSI}, 
    including implementation code, dependency and configuration files, data-preparation instructions, and scripts for the main experimental pipeline. The raw datasets remain available from their original public sources, and the repository documents data access, preprocessing, training, and inference procedures.

\item {\bf Experimental setting/details}
    \item[] Question: Does the paper specify all the training and test details (e.g., data splits, hyperparameters, how they were chosen, type of optimizer) necessary to understand the results?
    \item[] Answer: \answerYes{} 
    \item[] Justification: Training and test details are specified in \S\ref{sec:experiments} and Appendix ~\ref{app:impl}, including data splits, sensor budgets, optimizer and sampler settings, segment lengths, model ranks, and per-benchmark hyperparameters in Table 8; executable settings are provided in the public repository's configuration files.

\item {\bf Experiment statistical significance}
    \item[] Question: Does the paper report error bars suitably and correctly defined or other appropriate information about the statistical significance of the experiments?
    \item[] Answer: \answerYes{} 
    \item[] Justification: For each benchmark and sensor budget, we independently train five models with different random seeds and evaluate each trained model on the same held-out test set. Appendix D reports the mean $\pm$ standard deviation across the five independent training runs and paired $t$-tests on matched test-case results against the strongest baseline; the reported variability therefore captures training-run variability rather than only stochastic sampling at inference.

\item {\bf Experiments compute resources}
    \item[] Question: For each experiment, does the paper provide sufficient information on the computer resources (type of compute workers, memory, time of execution) needed to reproduce the experiments?
    \item[] Answer: \answerYes{} 
    \item[] Justification: Appendix ~\ref{app:impl_compute} explicitly states that all experiments were run on a single NVIDIA A100-80GB GPU, lists per-benchmark training times, and reports inference latency.
    
\item {\bf Code of ethics}
    \item[] Question: Does the research conducted in the paper conform, in every respect, with the NeurIPS Code of Ethics \url{https://neurips.cc/public/EthicsGuidelines}?
    \item[] Answer: \answerYes{} 
    \item[] Justification: The research uses public physics-sensing benchmarks and simulations (turbulent channel flow, GLORYS12 reanalysis, and OpenFOAM car aerodynamics on ShapeNet) and involves no human subjects, crowdsourcing, or sensitive personal data. Appendix F discusses potential societal impacts and dual-use considerations.

\item {\bf Broader impacts}
    \item[] Question: Does the paper discuss both potential positive societal impacts and negative societal impacts of the work performed?
    \item[] Answer: \answerYes{} 
    \item[] Justification: Appendix \ref{app:broimp} discusses potential benefits for climate monitoring, oceanographic surveys, and industrial diagnostics, including reduced sensing costs and environmental footprint. It also acknowledges potential dual-use concerns involving unauthorized tracking or surveillance and encourages responsible deployment.
    
\item {\bf Safeguards}
    \item[] Question: Does the paper describe safeguards that have been put in place for responsible release of data or models that have a high risk for misuse (e.g., pre-trained language models, image generators, or scraped datasets)?
    \item[] Answer: \answerNA{} 
    \item[] Justification: The paper introduces a physics-sensing reconstruction and trajectory optimization framework. It does not release pre-trained language models, image generators, or large-scale scraped datasets that carry high risk of misuse or dual-use, so safeguards are not applicable.

\item {\bf Licenses for existing assets}
    \item[] Question: Are the creators or original owners of assets (e.g., code, data, models), used in the paper, properly credited and are the license and terms of use explicitly mentioned and properly respected?
    \item[] Answer: \answerNo{} 
    \item[] Justification: The paper cites the original papers producing the datasets and software, and the public repository documents their sources. However, explicit version and license or terms-of-use information is not yet provided for every existing asset.

\item {\bf New assets}
    \item[] Question: Are new assets introduced in the paper well documented and is the documentation provided alongside the assets?
    \item[] Answer: \answerYes{} 
    \item[] Justification: The released code is documented in the public repository at \url{https://github.com/mengqi132/DyPSI}, which includes a README, dependency specifications, configuration files, usage instructions, and an MIT license. No new dataset or model checkpoint is released as part of this work.

\item {\bf Crowdsourcing and research with human subjects}
    \item[] Question: For crowdsourcing experiments and research with human subjects, does the paper include the full text of instructions given to participants and screenshots, if applicable, as well as details about compensation (if any)? 
    \item[] Answer: \answerNA{} 
    \item[] Justification: The paper does not involve crowdsourcing experiments or research with human subjects; all experiments are conducted on standard physics-sensing benchmarks and simulations.

\item {\bf Institutional review board (IRB) approvals or equivalent for research with human subjects}
    \item[] Question: Does the paper describe potential risks incurred by study participants, whether such risks were disclosed to the subjects, and whether Institutional Review Board (IRB) approvals (or an equivalent approval/review based on the requirements of your country or institution) were obtained?
    \item[] Answer: \answerNA{} 
    \item[] Justification: No human subjects are involved in this work; therefore, IRB approval or equivalent ethical review is not applicable.

\item {\bf Declaration of LLM usage}
    \item[] Question: Does the paper describe the usage of LLMs if it is an important, original, or non-standard component of the core methods in this research? Note that if the LLM is used only for writing, editing, or formatting purposes and does \emph{not} impact the core methodology, scientific rigor, or originality of the research, declaration is not required.
    \item[] Answer: \answerNA{} 
    \item[] Justification:  LLMs were not used as an important, original, or non-standard component of the core methodology, scientific rigor, or originality of this research. Any writing or formatting assistance did not impact the core technical contributions.

\end{enumerate}
\end{document}